\documentclass{article} 
\usepackage{iclr2026_conference,times}
\usepackage[utf8]{inputenc} 
\usepackage[T1]{fontenc}    
\usepackage{hyperref}       
\usepackage{url}            
\usepackage{booktabs}       
\usepackage{amsfonts}       
\usepackage{nicefrac}       
\usepackage{microtype}      
\usepackage{xcolor}         
\usepackage{amsmath}
\usepackage{amssymb}       
\usepackage{comment}
\usepackage{xspace}
\usepackage{marvosym}
\usepackage{booktabs}
\usepackage{tabularx}
\usepackage{array}
\usepackage{makecell}
\usepackage[table]{xcolor}
\usepackage{tikz}
\usepackage{wrapfig}
\usepackage{booktabs}

\definecolor{uclablue}{rgb}{0.15, 0.45, 0.68}

\definecolor{lightgreen}{RGB}{0,150,0}
\definecolor{myred}{RGB}{200,0,0} 

\usepackage{tabularx}
\robustify\bfseries
\usepackage{subcaption}
\usepackage{cleveref}

\definecolor{linkColor}{rgb}{0.2,0.4,0.6}
\definecolor{myblue}{HTML}{0379AC}
\definecolor{myred}{HTML}{A50E50}
\definecolor{myorange}{RGB}{238, 133, 74}
\definecolor{latentcolor}{named}{cyan}
\definecolor{normalcolor}{RGB}{0, 0, 0}
\usepackage{marvosym}
\usepackage{graphicx}     
\usepackage{fontawesome5}

\usepackage{amsmath,amsfonts,bm}

\def\eqref#1{equation~\ref{#1}}

\def\1{\bm{1}}

\DeclareMathAlphabet{\mathsfit}{\encodingdefault}{\sfdefault}{m}{sl}
\SetMathAlphabet{\mathsfit}{bold}{\encodingdefault}{\sfdefault}{bx}{n}

\usepackage{hyperref}
\usepackage{url}
\newcommand{\github}{\raisebox{-1.5pt}{\includegraphics[height=1.05em]{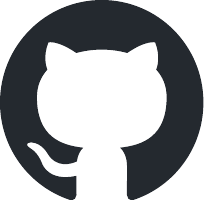}}\xspace}

\title{DelTA: Discriminative Token Credit \\ Assignment for Reinforcement Learning \\ from Verifiable Rewards}

\author{
  Kaiyi Zhang$^{1,2}$\hspace{0.4em}, \hspace{0.1em} Wei Wu$^{2}$\footnotemark[2]\hspace{0.4em}, \hspace{0.1em} Yankai Lin$^{1}$\footnotemark[2]\hspace{0.3em}\\
  $^{1}$Gaoling School of Artificial Intelligence, Renmin University of China \\ $^{2}$Ant International\\
  \Letter~\texttt{\{kyzhang, yankailin\}@ruc.edu.cn}\\
  \Letter~\texttt{wuwei19850318@gmail.com}  \\
  {\small \github \texttt{\textbf{Code:}} \url{https://github.com/RUCBM/DelTA}}\\
}

\iclrfinalcopy

\begin{document}
\renewcommand{\thefootnote}{\fnsymbol{footnote}}
\footnotetext[2]{Corresponding author: Wei Wu and Yankai Lin.}

\renewcommand{\thefootnote}{\arabic{footnote}}
\setcounter{footnote}{0}
\newcommand{\Eq}[1]{Eq.~(\ref{#1})}

\maketitle
\begin{abstract}
Reinforcement learning from verifiable rewards (RLVR) has emerged as a central technique for improving the reasoning capabilities of large language models. 
Despite its effectiveness, how response-level rewards translate into token-level probability changes remains poorly understood. 
We introduce a discriminator view of RLVR updates, showing that the policy-gradient update direction implicitly acts as a linear discriminator over token-gradient vectors and thereby determines which token probabilities are increased or decreased during learning. 
Under standard sequence-level RLVR, this discriminator is constructed from positive- and negative-side centroids formed by advantage-weighted averaging of token-gradient vectors. 
However, such centroid construction can be dominated by shared high-frequency patterns, such as formatting tokens, diluting sparse yet discriminative directions that better distinguish high-reward responses from low-reward ones.
To address this limitation, we propose \textbf{DelTA}, a discriminative token credit assignment method that estimates token coefficients to amplify side-specific token-gradient directions and downweight shared or weakly discriminative ones. These coefficients reweight a self-normalized RLVR surrogate, making the effective side-wise centroids more contrastive and thereby reshaping the RLVR update direction. 
On seven mathematical benchmarks, DelTA outperforms the strongest same-scale baselines by 3.26 and 2.62 average points on Qwen3-8B-Base and Qwen3-14B-Base, respectively. 
Additional results on code generation, a different backbone, and out-of-domain evaluations further demonstrate the generalization ability of DelTA.
\end{abstract}

\section{Introduction}

Reinforcement learning from verifiable rewards (RLVR) has become a key paradigm for improving the reasoning ability of large language models (LLMs), with strong gains in mathematics~\citep{shao2024deepseekmath,yang2024qwen2}, code generation~\citep{hui2024qwen2,shojaee2023execution,le2022coderl}, and formal problem solving~\citep{guo2025deepseek,team2025kimi}. 
RLVR optimizes response-level verifiable rewards, such as answer correctness, without requiring dense process-level annotations. 
This response-level supervision creates a granularity mismatch: each response provides a single scalar advantage, while the policy update is accumulated through token-level  terms. 
Recent studies show that RLVR induces sparse token-level distributional shifts, where substantial probability changes concentrate on a small subset of tokens while most token distributions change little~\citep{meng2026sparse,ma2026fipo}. 
This contrast suggests that sequence-level RLVR contains an implicit token-level selection mechanism that is not directly specified by the reward signal. 
Hence, an essential question arises: \textit{which token probabilities are increased or decreased by an RLVR update, and what determines these changes?}

We introduce a discriminator view of RLVR to explain this implicit token selection. 
Although an RLVR update is usually viewed as a parameter-space movement, the same update also defines a token-level decision rule: it determines whether a candidate-token probability is increased or decreased by the update.
The rule works by comparing token-gradient directions. 
For a sequence-level RLVR objective, the update direction contrasts token-gradient aggregates from positive-advantage responses and negative-advantage responses. 
After normalization, these aggregates define positive- and negative-side reference directions. 
A candidate-token probability is increased when its token-gradient vector aligns more with the positive-side reference direction than with the negative-side reference direction, and is decreased otherwise. In this sense, the RLVR update acts as an implicit linear discriminator over candidate token-gradient vectors. This view suggests that RLVR update directions can be understood and improved by analyzing and shaping the discriminator induced by the update.

Following the insights, further investigation indicates that standard sequence-level RLVR updates form the two-side directions by averaging token-gradient vectors from positive- and negative-advantage responses, yielding two centroids. 
Such centroids are natural summaries of each side, but a good within-side summary is not necessarily a good between-side discriminator. 
In reasoning tasks, higher- and lower-reward responses often share substantial common structure, such as formatting tokens or problem-specific entities. 
Because these shared patterns appear on both sides and occur frequently, their token-gradient directions can pull both centroids toward common background structure. 
Consequently, the induced discriminator may overemphasize task-agnostic commonalities and undermine sparse directions that better distinguish higher- from lower-reward responses.

To address this limitation, we propose \textbf{D}iscriminativ\textbf{e} signa\textbf{l}-guided \textbf{T}oken Credit \textbf{A}ssignment (\textbf{DelTA}). 
DelTA reshapes the induced RLVR discriminator by reweighting token-gradient terms in the RLVR surrogate. 
It estimates token coefficients that assign larger weights to token-gradient vectors more characteristic of their own advantage side than of the opposite side, while assigning smaller weights to shared or weakly discriminative directions. 
These coefficients change the effective aggregates that define the discriminator, making its positive and negative reference directions more contrastive and thereby reshaping the RLVR update direction. 
Empirically, DelTA consistently improves strong RLVR baselines. 
On seven mathematical benchmarks, it surpasses the strongest same-scale baseline by 3.26 average points on Qwen3-8B-Base and 2.62 points on Qwen3-14B-Base. 
It also improves code generation and generalizes to another backbone and out-of-domain evaluations.

In summary, our contributions are threefold. First, we introduce a local discriminator view of sequence-level RLVR, showing that policy-gradient updates induce an implicit linear discriminator over token-gradient vectors and thereby determine local token-probability changes. Second, using this view, we trace a limitation of standard sequence-level RLVR to the construction of the update direction: the side-wise centroids that form the induced discriminator can be pulled toward shared, high-frequency token-gradient directions, weakening its ability to separate token-gradient directions from higher- and lower-reward responses. Third, we propose DelTA, which reweights token-gradient terms by their positive-negative discriminative signal in a self-normalized RLVR surrogate, making the effective side-wise centroids more contrastive and consistently improving strong RLVR baselines across mathematical reasoning, code generation, different backbones, and out-of-domain evaluations.

\section{Preliminaries}
\label{sec:prelim}

We review the critic-free group-relative RLVR framework, taking DAPO as the main concrete example. For a prompt \(q\), let \(\{o_i\}_{i=1}^G\) be a group of sampled responses,
where \(o_i=(o_{i,1},\dots,o_{i,|o_i|})\). 
Each response receives a sequence-level reward \(R_i\). 
Let \(\mu_R\) and \(\sigma_R\) be the mean and standard deviation of rewards within the sampled group, and let \(\epsilon_A>0\) be a small numerical constant. 
The group-normalized advantage and token-level importance ratio are given by
\[
\hat A_i
=
\frac{R_i-\mu_R}{\sigma_R+\epsilon_A},
\qquad
r_{i,t}(\theta)
=
\frac{
\pi_\theta(o_{i,t}\mid q,o_{i,<t})
}{
\pi_{\theta_{\mathrm{old}}}(o_{i,t}\mid q,o_{i,<t})
}.
\]

\paragraph{DAPO-style Surrogate.}
DAPO~\citep{yu2025dapo} is a state-of-the-art critic-free group-relative RLVR method that optimizes a clipped surrogate objective with two key designs relevant to this work: asymmetric clipping and token-level normalization over all response tokens. The expected objective is defined as

\[
J_{\mathrm{DAPO}}(\theta)
=
\mathbb{E}\!\left[
\frac{1}{\sum_{i=1}^G |o_i|}
\sum_{i=1}^G \sum_{t=1}^{|o_i|}
\min\!\left(
r_{i,t}(\theta)\hat A_i,\;
\mathrm{clip}\!\big(
r_{i,t}(\theta),
1-\epsilon_{\mathrm{low}},
1+\epsilon_{\mathrm{high}}
\big)\hat A_i
\right)
\right].
\]
Here \(\hat A_i\) is defined at the response level and is therefore shared by all tokens in the same response. 
The per-token contribution to the surrogate is nevertheless accumulated through the token-level ratio \(r_{i,t}(\theta)\), which provides the basic object for the token-gradient analysis in the next section.

The original DAPO algorithm also includes dynamic sampling to encourage each sampled group to contain both correct and incorrect responses. 
This component affects rollout filtering rather than the form of the per-token surrogate above. 
We disable dynamic sampling for all methods in our experiments, and focus our analysis on the update rule induced by the surrogate objective.

\section{Method}
\label{sec:method}
\begin{figure*}[t]
\begin{center}
\centerline{\includegraphics[width=\linewidth]{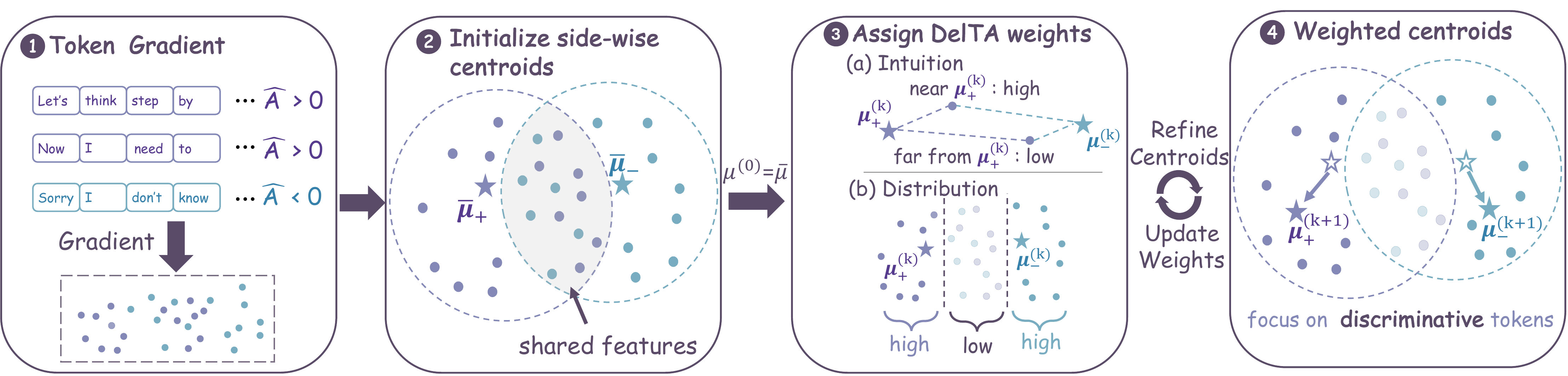}}
\caption{
Overview of DelTA. DelTA estimates token coefficients from the contrast between positive- and negative-advantage token-gradient aggregates, and uses the coefficients to reweight the sequence-level RLVR objective.
}
\label{fig:main}
\end{center}
\vskip -0.2in
\end{figure*}

Recent studies suggest that RLVR induces sparse and targeted changes at the token-distribution level: only a small fraction of token distributions undergo substantial shifts, while most remain nearly unchanged~\citep{meng2026sparse,ma2026fipo}. 
For sequence-level RLVR, this sparsity is not directly explained by the reward signal itself, since all tokens in the same response share the same scalar advantage. 
This suggests that the token-level selection effect is induced not by explicit token rewards, but by how token-gradient vectors are aggregated in the policy-gradient update.

We therefore analyze this induced selection effect by examining how sequence-level RLVR updates determine which candidate-token probabilities are increased or suppressed. Section~\ref{sec:motive} investigates this question using DAPO as a concrete instance and derives the discriminator view induced by RLVR updates. Although DAPO serves as the primary showcase, the resulting conclusions rely only on the update being expressible as an advantage-weighted aggregation of token-gradient vectors, and therefore naturally extend to a broader class of sequence-level RLVR objectives, as discussed in Appendix~\ref{app:centroid-view-rlvr}. Building on this analysis, Section~\ref{sec:DelTA} introduces DelTA, a new critic-free group-relative RLVR method that reweights token-gradient terms to reshape the induced discriminator and the corresponding RLVR update direction.

\subsection{A Local Discriminator View of RLVR Updates}
\label{sec:motive}

To understand how sequence-level RLVR implicitly selects tokens, we view the local policy update not only as a parameter update, but also as an implicit discriminator in token-gradient space.
For sequence-level RLVR objectives, the update direction contrasts token-gradient aggregates from positive- and negative-advantage responses.
After normalization, these aggregates define two side-wise reference directions.
A candidate token log-probability is locally increased when its token-gradient vector aligns more with the positive reference direction than with the negative one, and is decreased otherwise.

Formally, let \(c\) denote an arbitrary generation context, \(x\) a candidate next token under this context, and \(\pi_\theta(x\mid c)\) the policy model. 
For a local parameter update \(\Delta\theta\) around \(\theta_{\mathrm{old}}\), a first-order Taylor expansion gives
\begin{equation}
\label{eq:local-logprob-change}
\Delta \log \pi(x\mid c)
:=
\log \pi_{\theta_{\mathrm{old}}+\Delta\theta}(x\mid c)
-
\log \pi_{\theta_{\mathrm{old}}}(x\mid c)
\approx
\left(
\nabla_\theta \log \pi_\theta(x\mid c)
\Big|_{\theta=\theta_{\mathrm{old}}}
\right)^{\top}
\Delta\theta .
\end{equation}
Thus, once \(x\), \(c\), and \(\theta_{\mathrm{old}}\) are fixed, the local increase or decrease of the candidate token probability is determined by the inner product between its token-gradient vector and the update direction \(\Delta\theta\). In the following analysis, we focus on the local update direction (i.e., \(\Delta\theta\)).

For concreteness, consider the DAPO-style sequence-level surrogate used in our analysis. 
Let \(\{o_i\}_{i=1}^G\) be a rollout group, and let \(\hat A_i\) be the group-normalized advantage of response \(o_i\). 
Around \(\theta_{\mathrm{old}}\), clipping is locally inactive because \(r_{i,t}(\theta_{\mathrm{old}})=1\). 
The local policy-gradient update can therefore be written as an advantage-weighted aggregation of sampled-token gradients. 
Separating this aggregation by the sign of the response-level advantage gives
\begin{equation}
\label{eq:local-rlvr-update}
\Delta\theta_{\mathrm{RLVR}}
\propto
\sum_{i:\hat A_i>0}\sum_{t=1}^{|o_i|}
\hat A_i\,v_{i,t}
-
\sum_{i:\hat A_i<0}\sum_{t=1}^{|o_i|}
|\hat A_i|\,v_{i,t},
\qquad
v_{i,t}:=
\nabla_\theta
\log \pi_\theta(o_{i,t}\mid q,o_{i,<t})
\Big|_{\theta=\theta_{\mathrm{old}}}.
\end{equation}
We refer to token-gradient vectors from responses with \(\hat A_i>0\) as the \emph{positive side}, and those from responses with \(\hat A_i<0\) as the \emph{negative side}. 
Throughout this analysis, \(v_{i,t}\) denotes the exact full-parameter token-gradient vector. 
A detailed derivation of \Eq{eq:local-rlvr-update} from the DAPO surrogate is provided in Appendix~\ref{app:dapo-local-update}. We use this local characterization as an analysis and design principle for shaping the policy-update direction, rather than as an exact description of the full nonlinear clipped RLVR training trajectory.

\Eq{eq:local-rlvr-update} contains two components: the total mass of each side and the reference direction of each side. 
Let \(M_+=\sum_{i:\hat A_i>0}\sum_{t=1}^{|o_i|}\hat A_i\) and \(M_-=\sum_{i:\hat A_i<0}\sum_{t=1}^{|o_i|}|\hat A_i|\) denote the total positive and negative advantage masses. 
Then \Eq{eq:local-rlvr-update} can be rewritten as
\begin{equation}
\label{eq:centroid-decomposition}
\Delta\theta_{\mathrm{RLVR}}
\propto
M_+\bar\mu_+ - M_-\bar\mu_-,
\qquad
\bar\mu_+
=
\frac{
\sum_{i:\hat A_i>0}\sum_{t=1}^{|o_i|}
\hat A_i\,v_{i,t}
}{
M_+
},
\qquad
\bar\mu_-
=
\frac{
\sum_{i:\hat A_i<0}\sum_{t=1}^{|o_i|}
|\hat A_i|\,v_{i,t}
}{
M_-
}.
\end{equation}
Here, \(M_+\) and \(M_-\) determine the total strength of the two advantage sides, while \(\bar\mu_+\) and \(\bar\mu_-\) are their normalized aggregate directions. 
Substituting \Eq{eq:centroid-decomposition} into \Eq{eq:local-logprob-change} yields
\begin{equation}
\label{eq:local-discriminator-score}
\Delta \log \pi(x\mid c)
\propto
M_+
\left(
\nabla_\theta \log \pi_\theta(x\mid c)
\Big|_{\theta=\theta_{\mathrm{old}}}
\right)^{\top}
\bar\mu_+
-
M_-
\left(
\nabla_\theta \log \pi_\theta(x\mid c)
\Big|_{\theta=\theta_{\mathrm{old}}}
\right)^{\top}
\bar\mu_- .
\end{equation}
The two terms in \Eq{eq:local-discriminator-score} define the positive-side and negative-side scores assigned to the candidate token-gradient vector. 
The candidate token probability is locally increased when its positive-side score exceeds its negative-side score, and decreased otherwise. 
In this sense, the update direction has a dual role: in parameter space, it is a policy-update direction; in token-gradient space, it acts as an implicit linear discriminator over candidate token-gradient vectors. 
This discriminator is not explicitly parameterized or separately trained; it is induced by the policy-gradient update itself. 
This duality suggests a reverse design perspective: since the update direction induces a discriminator in token-gradient space, we can instead ask how to shape this induced discriminator and adjust the update direction accordingly. 
Thus, \textbf{RLVR update directions can be understood and improved by studying and shaping the local discriminator induced by the update}.

For the induced discriminator, the central objects are the side-wise reference directions \(\bar\mu_+\) and \(\bar\mu_-\). 
Under the standard sequence-level RLVR update, these directions are simply the advantage-weighted centroids of the token-gradient vectors on the positive and negative sides. 
Equivalently, they are weighted least-squares summaries that minimize within-side squared distances, as shown in Appendix~\ref{app:l2-centroid}. 
Such centroids are natural if the goal is to summarize each side independently. 
However, the induced discriminator uses them for a different purpose: distinguishing positive-advantage token gradients from negative-advantage token gradients.

This creates a mismatch between within-side summarization and between-side discrimination. 
In RLVR, positive- and negative-advantage responses often share frequent token patterns, such as common formatting tokens or problem-specific entities. 
The corresponding token-gradient directions can dominate both side-wise centroids, making the positive and negative reference directions less discriminative and diluting rarer directions that better separate higher-reward responses from lower-reward responses. 
From a classical discriminative perspective, good within-side summaries are not necessarily good between-side discriminators~\citep{cohen2013applied,zhao2024linear,khosla2020supervised}.

This motivates a centroid-level design principle: to obtain a better local update direction, we can reshape the side-wise centroids that define the induced discriminator. 
Changing these centroids changes the scores assigned to candidate token-gradient vectors, and therefore changes which token probabilities are locally increased or decreased. 
This principle motivates DelTA: we reshape the effective side-wise centroids by assigning larger weights to token-gradient directions that better distinguish the two advantage sides.

\subsection{DelTA: Discriminative Signal-guided Token Credit Assignment}
\label{sec:DelTA}

DelTA implements the centroid-level principle above by reweighting token terms in the RLVR surrogate. 
Since the side-wise centroids are induced by weighted token-gradient aggregation rather than separately parameterized, changing token weights directly reshapes these centroids, and hence the induced discriminator and local update direction. 
At a high level, DelTA has three steps. 
First, it initializes the positive and negative reference directions from the original advantage-weighted centroids. 
Second, it refines these directions through a small number of alternating steps: with the current centroids fixed, DelTA estimates discriminative token scores; with the scores fixed, it recomputes each side-wise centroid as a score-weighted average of token-gradient vectors from that side. 
Third, it maps the final scores to bounded coefficients and uses them to reweight the sequence-level RLVR surrogate.

The formulation below is written in terms of token-gradient vectors \(v_{i,t}\). 
In the exact version, these vectors are the full-parameter gradients defined in Section~\ref{sec:motive}. 
In practice, explicitly materializing full-parameter gradients for all rollout tokens is computationally prohibitive at LLM scale, so we use a layer-restricted LM-head gradient representation to compute the stop-gradient token coefficients. 
This proxy affects only the coefficient computation; the analysis remains formulated with full-parameter token gradients, and the resulting weighted RLVR objective is still optimized over the full policy parameters. 
Further details and proxy ablations are provided in Appendix~\ref{app:proxy}.

We initialize the refinement from the original advantage-weighted centroids, \(\mu_+^{(0)}=\bar\mu_+\) and \(\mu_-^{(0)}=\bar\mu_-\). 
DelTA then runs \(K\) stop-gradient refinement iterations. 
At iteration \(k=0,\dots,K-1\), DelTA first estimates a soft discriminative score \(\alpha_{i,t}^{(k)}\) for each token-gradient vector. 
We describe the positive side; the negative side is obtained symmetrically by swapping \(\mu_+^{(k)}\) and \(\mu_-^{(k)}\), and by replacing \(\gamma_+^{(k)}\) with \(\gamma_-^{(k)}\). 
For a positive-advantage token, DelTA assigns a larger score when \(v_{i,t}\) is closer to the positive centroid than to the negative centroid. 
Specifically, \(\alpha_{i,t}^{(k)}\) is defined by the entropy-regularized assignment problem
\begin{equation}
\label{eq:DelTA-assignment-objective}
\alpha_{i,t}^{(k)}
=
\arg\max_{\alpha\in[0,1]}
\alpha
\Big(
\|v_{i,t}-\mu_-^{(k)}\|_2^2
-
\|v_{i,t}-\mu_+^{(k)}\|_2^2
\Big)
+
\gamma_+^{(k)}h(\alpha),
\qquad \hat A_i>0,
\end{equation}
where \(h(\alpha)=-\alpha\log\alpha-(1-\alpha)\log(1-\alpha)\) is the binary entropy regularizer, and \(\gamma_+^{(k)}>0\) is a side-specific temperature for the positive-side assignment. 
The distance-margin term is positive when \(v_{i,t}\) is closer to the positive centroid than to the negative centroid, so maximizing \Eq{eq:DelTA-assignment-objective} assigns a larger score to tokens that are more characteristic of their own side. 
The entropy regularizer and the temperature jointly control the softness of this assignment: smaller temperatures make the score closer to a hard decision, while larger temperatures produce smoother scores. 
We use squared Euclidean distances to stay consistent with the centroid construction, as detailed in Appendix~\ref{app:l2-centroid}.

For fixed centroids and temperatures, the closed-form solution is
\begin{equation}
\label{eq:DelTA-alpha}
\alpha_{i,t}^{(k)}
=
\begin{cases}
\sigma\!\left(
\dfrac{
\|v_{i,t}-\mu_-^{(k)}\|_2^2
-
\|v_{i,t}-\mu_+^{(k)}\|_2^2
}{
\gamma_+^{(k)}
}
\right),
& \hat A_i>0,\\[1.2ex]
\sigma\!\left(
\dfrac{
\|v_{i,t}-\mu_+^{(k)}\|_2^2
-
\|v_{i,t}-\mu_-^{(k)}\|_2^2
}{
\gamma_-^{(k)}
}
\right),
& \hat A_i<0,
\end{cases}
\end{equation}
where \(\sigma(\cdot)\) is the sigmoid function. 
The side-specific temperatures \(\gamma_+^{(k)}\) and \(\gamma_-^{(k)}\) adapt the assignment scale for the two advantage sides; their computation is detailed in Appendix~\ref{app:DelTA-iteration}. 
A derivation of \Eq{eq:DelTA-alpha} is provided in Appendix~\ref{app:DelTA-weight-derivation}. 

Thus, \(\alpha_{i,t}^{(k)}\) is large when the token-gradient vector is more representative of its own advantage side than of the opposite side, and small for shared or weakly discriminative directions. 

Given these scores, DelTA updates the centroids as score-weighted within-side averages:
\begin{equation}
\label{eq:DelTA-center-update}
\mu_+^{(k+1)}
=
\frac{
\sum_{i:\hat A_i>0}\sum_{t=1}^{|o_i|}
\hat A_i\,\alpha_{i,t}^{(k)}\,v_{i,t}
}{
\sum_{i:\hat A_i>0}\sum_{t=1}^{|o_i|}
\hat A_i\,\alpha_{i,t}^{(k)}
},
\qquad
\mu_-^{(k+1)}
=
\frac{
\sum_{i:\hat A_i<0}\sum_{t=1}^{|o_i|}
|\hat A_i|\,\alpha_{i,t}^{(k)}\,v_{i,t}
}{
\sum_{i:\hat A_i<0}\sum_{t=1}^{|o_i|}
|\hat A_i|\,\alpha_{i,t}^{(k)}
}.
\end{equation}
This refinement gives larger influence to token-gradient vectors that are more characteristic of their own side, while downweighting shared or weakly discriminative directions. 
It is used only to compute stop-gradient token coefficients; no gradients are propagated through the refinement, and no additional loss term is added.

After the final refinement step, DelTA recomputes raw scores \(\alpha_{i,t}^{\star}\) with the refined centroids and maps them to bounded coefficients \(\lambda_{i,t}=\lambda_{\min}+(\lambda_{\max}-\lambda_{\min})\alpha_{i,t}^{\star}\). 
The bounded range prevents extreme reweighting while preserving the ranking of the discriminative scores. 
DelTA then replaces the uniform token average in DAPO with the following self-normalized weighted surrogate:
\begin{equation}
\label{eq:DelTA-weighted-objective}
J_{\mathrm{DelTA}}(\theta)
=
\mathbb{E}\!\left[
\frac{
1
}{
\sum_{i=1}^{G}\sum_{t=1}^{|o_i|}\lambda_{i,t}
}
\sum_{i=1}^{G}\sum_{t=1}^{|o_i|}
\lambda_{i,t}
\min\!\Big(
r_{i,t}(\theta)\hat A_i,\;
\mathrm{clip}\!\big(
r_{i,t}(\theta),
1-\epsilon_{\mathrm{low}},
1+\epsilon_{\mathrm{high}}
\big)\hat A_i
\Big)
\right].
\end{equation}

Around \(\theta_{\mathrm{old}}\), \Eq{eq:DelTA-weighted-objective} changes each sampled-token contribution from \(\hat A_i v_{i,t}\) to \(\lambda_{i,t}\hat A_i v_{i,t}\). 
This reweighting reshapes the effective side-wise centroids, and hence the induced discriminator and local RLVR update direction, by amplifying side-specific token-gradient directions and downweighting shared or weakly discriminative ones. 
The coefficients are stop-gradient quantities computed once per rollout batch, fixed across optimization epochs, and recomputed after new trajectories are sampled. 
Full details are provided in Appendix~\ref{app:DelTA-iteration}.

\section{Experiments}
\subsection{Experimental Setup}
\label{sec:exp_s}
We train on two backbones, Qwen3-8B-Base~\citep{yang2025qwen3} and Qwen3-14B-Base, using DeepMath-103K~\citep{he2025deepmath} with VeRL~\citep{sheng2024hybridflow}. For DelTA, we set \([\lambda_{\min},\lambda_{\max}]=[0.8,1.2]\) and use one refinement iteration (\(K=1\)). We compare against DAPO~\citep{yu2025dapo}, DAPO w/ Forking Tokens (DAPO w/ FT)~\citep{wang2025beyond}, SAPO~\citep{gao2025soft}, and FIPO~\citep{ma2026fipo}, training all methods with the same hyperparameters. We disable dynamic sampling for all methods to isolate the effect of the policy-update objective. Detailed training settings and baseline descriptions are provided in Appendix~\ref{app:hyp} and Appendix~\ref{app:baseline-details}, respectively.

We evaluate our models on seven mathematical benchmarks: AIME24~\citep{aime24}, AIME25~\citep{aime25}, AIME26~\citep{aime26}, HMMT25 (February)~\citep{balunovic_srimatharena_2025}, HMMT25 (November), HMMT26 (February) and Brumo 25.  To better reveal each model's long-reasoning capability, we set the maximum generation length during evaluation to 30{,}000 tokens. We sample 16 responses for each problem. We report the average performance over all samples. Unless otherwise specified, Avg. denotes the question-count weighted average across benchmarks. Detailed hyperparameters are provided in Appendix~\ref{app:hyp}.

\subsection{Main Results}
\label{sec:main-results}
\newcommand{\bodystrut}{\rule[-0.25ex]{0pt}{2.45ex}}

\begin{table*}[t]
\centering
\caption{Main results on seven mathematical reasoning benchmarks for Qwen3-8B-Base and Qwen3-14B-Base. DelTA consistently outperforms all compared same-scale RL baselines. The best results are in \textbf{bold}, and the second-best results are \underline{underlined}.}
\label{tab:results_main}
\small
\setlength{\tabcolsep}{2pt}
\renewcommand{\arraystretch}{1.06}

\begin{tabularx}{\textwidth}{@{}>{\raggedright\arraybackslash}p{0.16\textwidth}*{8}{>{\centering\arraybackslash}X}@{}}
\toprule
\textbf{Method}
& AIME24
& AIME25
& AIME26
& \makecell[c]{HMMT25\\(Feb.)}
& \makecell[c]{HMMT25\\(Nov.)}
& \makecell[c]{HMMT26\\(Feb.)}
& Brumo25
& Avg. \\
\midrule

\multicolumn{9}{@{}l@{}}{\emph{\textbf{Qwen3-8B-Base}}} \\
\bodystrut DAPO        & 34.79 & 23.33 & 24.17 & 13.54 & 12.08 & 16.86 & 36.46 & 22.95 \\
\bodystrut DAPO w/ FT  & 36.67 & 23.96 & \underline{26.46} & \underline{15.62} & 15.42 & 17.05 & 39.17 & 24.80 \\
\bodystrut SAPO        & \underline{38.75} & \underline{24.37} & 26.25 & 14.58 & \underline{16.04} & 17.42 & \underline{39.37} & \underline{25.14} \\
\bodystrut FIPO        & 37.50 & 23.13 & 23.96 & 14.58     & 12.92 & \underline{17.99} & 37.71     & 23.89 \\
\rowcolor{cyan!10}
\bodystrut \textbf{DelTA} & \textbf{43.13} & \textbf{26.46} & \textbf{28.12} & \textbf{18.33}
              & \textbf{18.54} & \textbf{20.27} & \textbf{44.79} & \textbf{28.40} \\
\midrule

\multicolumn{9}{@{}l@{}}{\emph{\textbf{Qwen3-14B-Base}}} \\
\bodystrut DAPO        & 51.25 & 32.29 & 39.79 & 19.79 & 30.00 & 25.38 & 48.13 & 35.09 \\
\bodystrut DAPO w/ FT  & 54.37 & 33.75 & 41.46 & 20.42 & 31.67 & \underline{24.81} & 52.08 & 36.77 \\
\bodystrut SAPO        & 53.96 & 34.17 & 41.46 & 20.62 & 28.33 & 24.05 & 50.21 & 35.94 \\
\bodystrut FIPO        & \underline{54.58} & \underline{35.00} & \underline{42.50} & \underline{21.46} & \underline{32.29} & 24.43 & \underline{52.08} & \underline{37.29} \\
\rowcolor{cyan!10}
\bodystrut \textbf{DelTA} & \textbf{56.87} & \textbf{37.92} & \textbf{45.21} & \textbf{26.04}
              & \textbf{32.92} & \textbf{26.89} & \textbf{54.79} & \textbf{39.91} \\
\bottomrule
\end{tabularx}
\end{table*}
Table~\ref{tab:results_main} reports the main results on seven mathematical
reasoning benchmarks. DelTA consistently outperforms all same-scale RL baselines
on both Qwen3-8B-Base and Qwen3-14B-Base, achieving the best result on every
benchmark and the highest average score at both scales. Compared with the
strongest same-scale baseline, DelTA improves the average score from \(25.14\)
to \(28.40\) on the 8B backbone and from \(37.29\) to \(39.91\) on the 14B
backbone. As detailed in Appendix~\ref{app:sig}, under repeated-generation evaluation, DelTA significantly outperforms the strongest same-scale baselines.

The consistent gains across benchmarks and model scales suggest that DelTA improves the policy-update mechanism beyond a single benchmark instance. The computational overhead of DelTA is discussed in Appendix~\ref{app:expense}. We further evaluate DelTA beyond the main math setting: Appendix~\ref{app:code-gen} shows that DelTA also improves DAPO on code generation benchmarks, and Appendix~\ref{app:other-architectures} shows that DelTA remains effective on Olmo3-7B-Base~\citep{olmo2025olmo}.  These results indicate that the benefit of DelTA is not limited to a specific benchmark family or backbone architecture.

\subsection{Training Dynamics}
\label{sec:training-dynamics}
\begin{figure*}[h]
    \centering
    \begin{minipage}{0.32\textwidth}
        \centering
        \includegraphics[width=\textwidth]{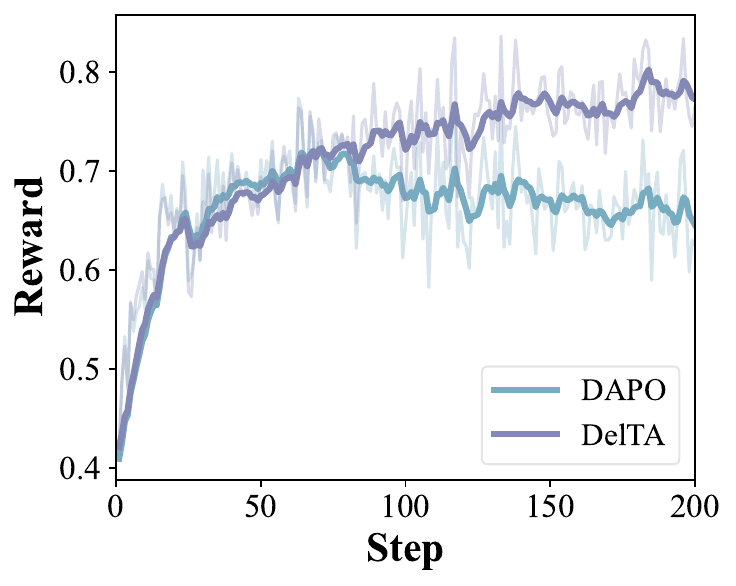}
        \captionsetup{labelformat=empty}
    \end{minipage}
    \hfill
    \begin{minipage}{0.32\textwidth}
        \centering
        \includegraphics[width=\textwidth]{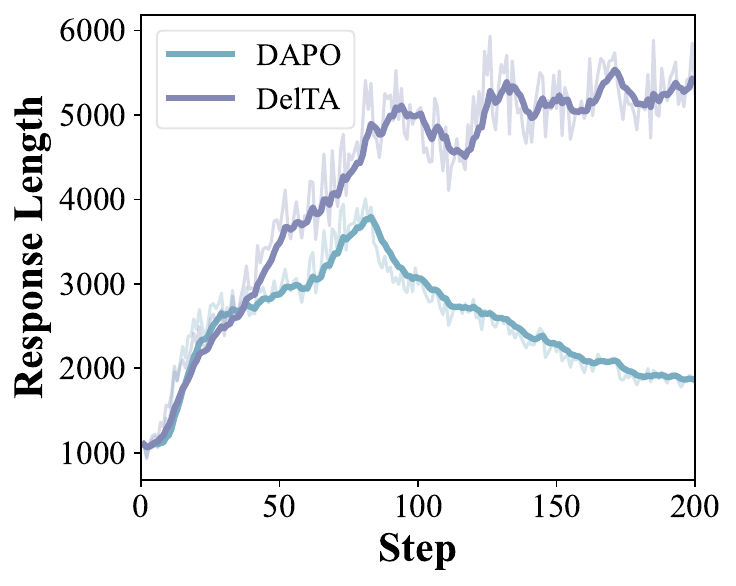}
        \captionsetup{labelformat=empty}
    \end{minipage}
    \hfill
    \begin{minipage}{0.32\textwidth}
        \centering
        \includegraphics[width=\textwidth]{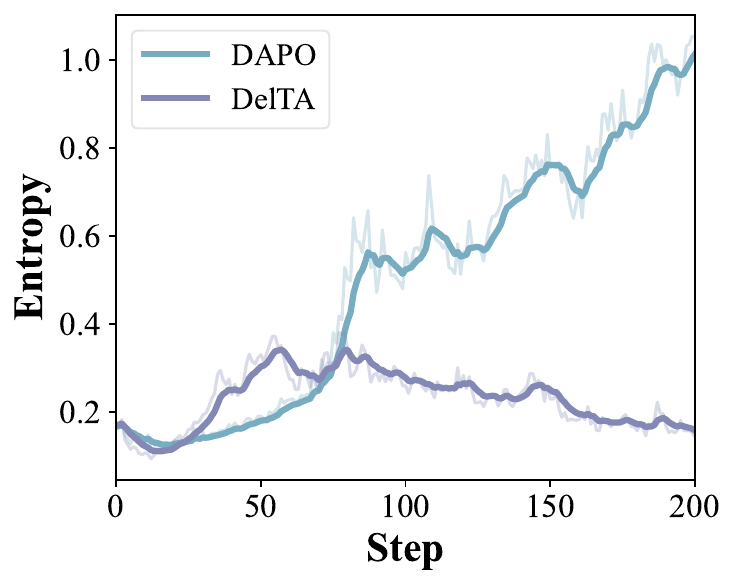}
        \captionsetup{labelformat=empty}
    \end{minipage}
    \caption{Training dynamics of DelTA~compared with DAPO. \textbf{Left:} Reward; \textbf{Middle:} Response Length; \textbf{Right:} Entropy.}
    \label{fig:training_dynamics}
\end{figure*}

Figure~\ref{fig:training_dynamics} compares the training dynamics of DelTA and DAPO on Qwen3-8B-Base. 
The two methods show similar reward trajectories in the early stage, but diverge afterwards: DAPO plateaus and slightly degrades, whereas DelTA continues to improve and reaches a higher final reward. 
The response-length and entropy curves suggest that this divergence is not merely due to shorter answers. 
DAPO shifts toward shorter responses with rising entropy, whereas DelTA maintains longer responses with lower entropy and higher reward, indicating more stable and confident long-reasoning behavior.

This behavior is consistent with our discriminator view. 
In standard sequence-level aggregation, shared background directions can dominate the side-wise centroids and weaken the contrast of the induced update direction. 
DelTA counteracts this by upweighting side-specific token-gradient directions and downweighting shared or weakly discriminative ones. 
As a result, the effective reference directions become more contrastive, helping the update sustain useful long-reasoning trajectories without an explicit length incentive.

\section{Analysis}
In this section, we provide further analysis of DelTA beyond the main results.
Unless stated otherwise, we analyze DelTA on Qwen3-8B-Base using four representative math benchmarks: AIME25, AIME26, HMMT25-Nov, and HMMT26-Feb, abbreviated as HMMT25 and HMMT26.

We organize the analysis around five diagnostic questions:
\textbf{Q1:} Is the opposite-side comparison necessary?
Section~\ref{sec:own-side-main} tests whether own-side centrality alone can explain DelTA's gains.
\textbf{Q2:} Does $\lambda_{i,t}$ capture useful
token-level learning signals?
Section~\ref{sec:lambda-effectiveness} examines this by using $\lambda_{i,t}$
only for token selection.
\textbf{Q3:} Are the design components of DelTA necessary?
Section~\ref{sec:ablation} answers this through component ablations.
\textbf{Q4:} Is DelTA sensitive to its hyperparameters?
Appendix~\ref{app:hyp-sensitivity} studies the robustness of DelTA
under different hyperparameter choices.
\textbf{Q5:} Does DelTA generalize to out-of-domain evaluation?
Appendix~\ref{app:ood} evaluates DelTA on additional OOD benchmarks.

\subsection{Q1: Is the opposite-side comparison necessary?}
\label{sec:own-side-main}

To test whether DelTA's gain can be explained by own-side centrality alone, we construct a within-side-only variant.
This variant keeps the same coefficient normalization and weighted DAPO objective as DelTA, but removes the opposite-side distance from the assignment score. For tokens from positive-advantage responses, we use
$
\alpha_{i,t}
=
\sigma\!\left(
-\frac{\|v_{i,t}-\mu_+\|_2^2}{\gamma_+}
\right),
$
and define the negative side symmetrically with \(\mu_-\) and \(\gamma_-\).
Thus, this variant assigns larger coefficients to tokens closer to their own-side centroid, without considering their distance to the opposite-side centroid.
\begin{wraptable}{r}{0.65\textwidth}
\centering
\caption{
Effect of using only within-side information. The best results are in \textbf{bold}.
}
\label{tab:ana-own}
\small
\setlength{\tabcolsep}{4pt}
\renewcommand{\arraystretch}{1.06}

\begin{tabularx}{\linewidth}{@{}l*{5}{>{\centering\arraybackslash}X}@{}}
\toprule
\textbf{Method}
& AIME25
& AIME26
& HMMT25
& HMMT26
& Avg. \\
\midrule

\rowcolor{cyan!10}
\bodystrut \textbf{DelTA}
& \textbf{26.46} & \textbf{28.12} & \textbf{18.54} & \textbf{20.27} & \textbf{23.27} \\

\midrule
\bodystrut DAPO 
& 23.33  & 24.17 & 12.08 & 16.86 & 19.05 \\
\bodystrut Within-side only
& 21.67 & 22.08 & 11.04 & 17.05 & 17.94 \\

\bottomrule
\end{tabularx}
\vspace{-5pt}
\end{wraptable}
Table~\ref{tab:ana-own} shows that the within-side-only variant performs worse than both DelTA and the DAPO baseline. This result indicates that DelTA's gains cannot be explained by simply assigning larger weights to tokens close to their own-side centroid. In fact, own-side centrality alone can be misleading: tokens near a side-wise centroid may correspond to shared patterns rather than directions that distinguish positive- from negative-advantage responses. The opposite-side comparison is therefore essential, because it assigns high coefficients only to directions that are relatively more representative of their own side than of the opposite side.

\subsection{Q2: Does \texorpdfstring{\(\lambda_{i,t}\)}{lambda} capture useful token-level learning signals?}

\label{sec:lambda-effectiveness}

\begin{figure*}[t]
    \centering
    \begin{minipage}[t]{0.48\textwidth}
        \centering
        \includegraphics[width=\textwidth]{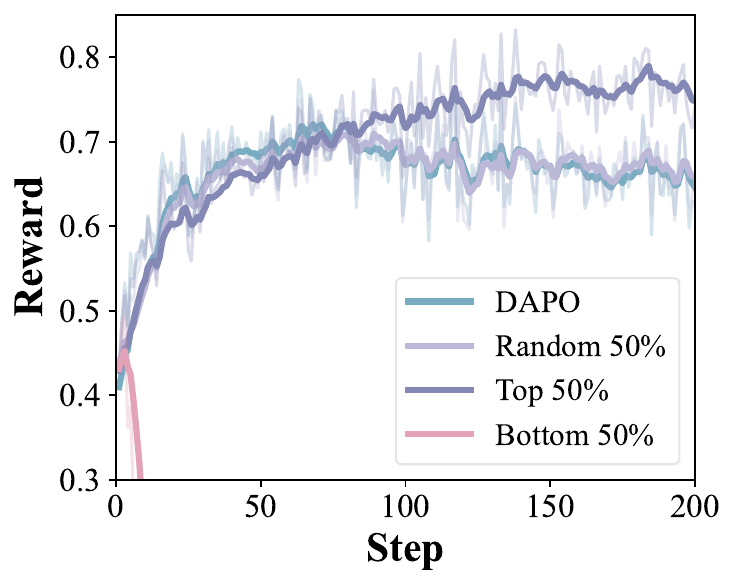}
        \captionof{figure}{Training reward under different token-selection strategies.}
        \label{fig:reward_mask}
    \end{minipage}
    \hfill
    \begin{minipage}[t]{0.48\textwidth}
        \centering
        \includegraphics[width=\textwidth]{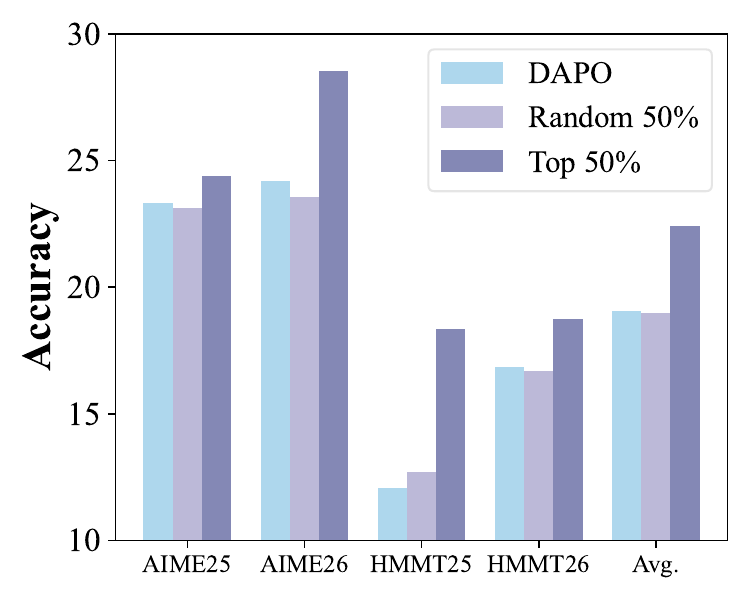}
        \captionof{figure}{Evaluation accuracy under different token-selection strategies.}
        \label{fig:acc_mask}
    \end{minipage}
    \vspace{-2mm}
\end{figure*}

We next test whether the DelTA coefficient \(\lambda_{i,t}\) identifies token-gradient directions with useful learning value.
To isolate the ranking effect from the continuous reweighting objective, we use \(\lambda_{i,t}\) only for hard token selection.
Specifically, we compute DelTA coefficients for all valid response tokens and train DAPO using only the top \(50\%\) tokens ranked by \(\lambda_{i,t}\).
As controls, we train on a random \(50\%\) subset and on the bottom \(50\%\) tokens, while keeping the standard DAPO loss unchanged.

Figure~\ref{fig:reward_mask} shows a sharp separation between these selections.
Training on the top-\(\lambda\) tokens consistently outperforms both full-token DAPO and random \(50\%\) selection, even though it uses only half of the token-gradient terms.
In contrast, training on the bottom-\(\lambda\) tokens quickly collapses.
The evaluation results in Figure~\ref{fig:acc_mask} show the same trend: top-\(\lambda\) training improves accuracy across benchmarks, whereas random selection remains close to DAPO.
These results indicate that DelTA's coefficients capture more than a benign sparsification signal.
If the gain came merely from reducing the number of optimized tokens, random \(50\%\) selection should provide a similar benefit, which it does not.
Instead, the fact that the top half outperforms full-token DAPO while the bottom half collapses suggests that low-\(\lambda\) tokens are not simply uninformative; their gradient directions can actively harm the RLVR update.
Thus, \(\lambda_{i,t}\) separates token-gradient directions with high effective learning value from shared or misleading directions that weaken policy improvement.

\subsection{Q3: Are the design components of DelTA necessary?}
\label{sec:ablation}

We ablate five design choices in DelTA while keeping all other training settings unchanged:
\textbf{w/o adaptive \(\gamma\)} fixes the assignment temperatures to the initial distance scale;
\textbf{w/o \(h(\alpha)\)} removes the entropy regularizer and turns the soft assignment into a hard \(0/1\) decision;
\textbf{w/o \(\lambda\)-norm} keeps the token coefficients in the numerator, but replaces DelTA's coefficient-mass normalizer \(1/\sum_{i,t}\lambda_{i,t}\) with the standard DAPO token-count normalizer \(1/\sum_i |o_i|\); \textbf{w/o range map} removes the linear mapping from assignment scores to \([\lambda_{\min},\lambda_{\max}]\), and directly uses the raw scores \(\alpha_{i,t}\in(0,1)\) as token weights; and \textbf{w/o refinement} estimates coefficients only from the initial side-wise centroids.

\begin{wraptable}{r}{0.65\textwidth}
\centering
\caption{
Ablation study of DelTA. The best results are in \textbf{bold}, and the second-best results are \underline{underlined}.
}
\label{tab:ablation}
\small
\setlength{\tabcolsep}{4pt}
\renewcommand{\arraystretch}{1.06}

\begin{tabularx}{\linewidth}{@{}l*{5}{>{\centering\arraybackslash}X}@{}}
\toprule
\textbf{Method}
& AIME25
& AIME26
& HMMT25
& HMMT26
& Avg. \\
\midrule

\rowcolor{cyan!10}
\bodystrut \textbf{Full DelTA}
& \textbf{26.46} & \textbf{28.12} & \textbf{18.54} & \textbf{20.27} & \textbf{23.27} \\

\midrule
\bodystrut w/o adaptive \(\gamma\)
& \underline{25.00} & 26.04 & \underline{16.04} & 17.99 & 21.19 \\
\bodystrut w/o \(h(\alpha)\)
& 24.37 & \underline{26.87} & 15.42 & 17.42 & 20.93 \\
\bodystrut w/o \(\lambda\)-norm
& 24.37 & 26.25 & 15.83 & \underline{19.32} & \underline{21.39} \\
\bodystrut w/o range map
& 24.79 & 25.83 & 15.83 & 17.05 & 20.78 \\
\bodystrut w/o refinement
& 23.13 & 25.42 & 15.42 & 16.29 & 19.97 \\

\bottomrule
\end{tabularx}
\vspace{-10pt}
\end{wraptable}

Table~\ref{tab:ablation} shows that each component contributes to DelTA.
Removing any design choice reduces the average score.
The largest drop comes from \textbf{w/o refinement}, indicating that one-shot coefficients from the initial centroids are insufficient.
The drops from \textbf{w/o range map} and \textbf{w/o \(h(\alpha)\)} suggest that bounded soft coefficients are more stable than raw scores or hard assignments.
Finally, the degradation from \textbf{w/o adaptive \(\gamma\)} and \textbf{w/o \(\lambda\)-norm} shows that scale adaptation and coefficient-mass normalization are both useful.

\section{Related Work}

Reinforcement learning has become an effective paradigm for improving LLM reasoning, especially in domains with verifiable feedback such as mathematics. Representative methods include PPO-style optimization and recent critic-free group-relative objectives such as GRPO and DAPO~\citep{schulman2017proximal,shao2024deepseekmath,yu2025dapo}. Recent work further studies the mechanisms of RLVR~\citep{yue2025does,huan2025does,meng2026sparse}, improves training stability and efficiency~\citep{zheng2025group,gao2025soft,liu2025understanding}, and explores off-policy or semi-off-policy training~\citep{yan2025learning,zhang2025stephint}. 

A central challenge in RLVR is that rewards are usually provided at the response level, while policy updates are applied at the token level. Prior work addresses this mismatch through token-level or step-level reweighting~\citep{kazemnejad2025vinepporefiningcreditassignment,xie2025capo}, process reward models or learned value functions~\citep{cui2025process,zhang2025lessons}, and token-selection signals such as entropy or future influence~\citep{wang2025beyond,ma2026fipo}. 

Our work is complementary. Instead of relying on external dense rewards, value estimates, or auxiliary token-selection rules, DelTA reweights the RLVR surrogate using token coefficients derived from the positive-negative discriminator induced by the update itself.

\section{Conclusion}
We studied token-level learning in RLVR from a local discriminative perspective.
Using DAPO as a representative sequence-level objective, we showed that its policy-gradient update induces an implicit linear discriminator over candidate token-gradient vectors.
This discriminator is defined by positive and negative side-wise centroids, which can be dominated by shared, weakly discriminative directions rather than directions that distinguish higher- from lower-reward responses.
Motivated by the mismatch between within-side summarization and between-side discrimination, we proposed \textbf{DelTA}, which estimates token coefficients from refined centroid contrasts and uses them to reweight a self-normalized clipped RLVR surrogate.
Experiments on mathematical reasoning, code generation, another model family, and out-of-domain benchmarks show consistent improvements over strong RLVR baselines.
Overall, our results suggest that shaping the discriminator induced by RLVR updates is a useful route for improving token-level credit assignment under sequence-level supervision.

\bibliography{iclr2026_conference}
\bibliographystyle{iclr2026_conference}

\newpage
\appendix
\section{Limitations}
DelTA provides a lightweight way to incorporate discriminative token-level credit assignment into sequence-level RLVR. 
While our experiments show consistent gains across several settings, there remain several directions for further improvement and broader validation.

First, DelTA estimates token coefficients using a layer-restricted token-gradient proxy rather than full-parameter token gradients, since computing full gradients for all sampled tokens is computationally expensive at RLVR scale. 
This proxy is used only for stop-gradient coefficient estimation, and the weighted RLVR objective still updates the full policy parameters. 
Our proxy ablations show that DelTA is robust to different layer-restricted proxy choices, but exploring richer and more efficient token-gradient approximations is a promising direction for future work.

Second, our empirical evaluation focuses primarily on mathematical reasoning, with additional validation on code generation, different backbone architectures, and out-of-domain benchmarks. 
Future work could further evaluate DelTA on broader RLVR settings, including multi-turn interaction, tool-use tasks, and domains with more diverse verifiable signals. 

Finally, DelTA introduces additional computation for coefficient estimation. 
As discussed in Appendix~\ref{app:expense}, the measured overhead is modest in our setting, and future engineering improvements such as more efficient caching or lower-cost proxy computation could further reduce this cost.

\section{Broader impacts}
This work studies token-level credit assignment for reinforcement learning from verifiable rewards. 
Its potential positive impact is to improve the effectiveness and efficiency of training reasoning-capable language models, especially in domains where correctness can be automatically verified, such as mathematics and code generation. 
By providing a more interpretable view of how sequence-level RLVR updates allocate token-level credit, the work may also help researchers better diagnose and improve RL training dynamics.

At the same time, stronger reasoning models may also be misused in settings such as automated generation of misleading content, scalable code generation for harmful purposes, or other dual-use applications. 
DelTA does not introduce new user-facing deployment mechanisms, new datasets containing sensitive information, or new privacy-invasive capabilities, but it may improve the capabilities of models trained with RLVR. 
Responsible deployment of models trained with such methods should therefore follow standard safety practices, including appropriate evaluation, monitoring, access control when needed, and task-specific safeguards for high-risk applications.

\section{Derivation of the Local DAPO Update}
\label{app:dapo-local-update}

In this appendix, we derive the local first-order update form used in
Section~\ref{sec:motive}. We focus on the update rule induced by the DAPO
surrogate. Since dynamic sampling is disabled in our training recipe, it is not
included in the derivation.

For a fixed prompt \(q\), let \(\{o_i\}_{i=1}^G\) denote a sampled group of
responses drawn from the old policy \(\pi_{\theta_{\mathrm{old}}}\), where
\(o_i=(o_{i,1},\dots,o_{i,|o_i|})\). Let
\[
N := \sum_{i=1}^G |o_i|
\]
be the total number of valid response tokens in the group. Conditioning on this
rollout batch, the DAPO surrogate is
\[
J_{\mathrm{DAPO}}(\theta)
=
\frac{1}{N}
\sum_{i=1}^G \sum_{t=1}^{|o_i|}
\min\!\left(
r_{i,t}(\theta)\hat A_i,\;
\mathrm{clip}\!\big(
r_{i,t}(\theta),
1-\epsilon_{\mathrm{low}},
1+\epsilon_{\mathrm{high}}
\big)\hat A_i
\right),
\]
where
\[
r_{i,t}(\theta)
=
\frac{
\pi_\theta(o_{i,t}\mid q,o_{i,<t})
}{
\pi_{\theta_{\mathrm{old}}}(o_{i,t}\mid q,o_{i,<t})
}.
\]

We analyze a local update around \(\theta_{\mathrm{old}}\). At this point,
\(r_{i,t}(\theta_{\mathrm{old}})=1\) for every sampled token. Since \(1\) lies
inside the clipping interval, clipping is locally inactive, and the local
gradient of \(J_{\mathrm{DAPO}}\) agrees with that of the unclipped surrogate:
\[
\frac{1}{N}
\sum_{i=1}^G \sum_{t=1}^{|o_i|}
r_{i,t}(\theta)\hat A_i.
\]
Therefore,
\[
\nabla_\theta J_{\mathrm{DAPO}}(\theta)\Big|_{\theta=\theta_{\mathrm{old}}}
=
\frac{1}{N}
\sum_{i=1}^G \sum_{t=1}^{|o_i|}
\hat A_i
\nabla_\theta r_{i,t}(\theta)\Big|_{\theta=\theta_{\mathrm{old}}}.
\]

Because the denominator of \(r_{i,t}(\theta)\) is fixed with respect to
\(\theta\),
\[
\nabla_\theta r_{i,t}(\theta)\Big|_{\theta=\theta_{\mathrm{old}}}
=
\frac{
\nabla_\theta \pi_\theta(o_{i,t}\mid q,o_{i,<t})
\big|_{\theta=\theta_{\mathrm{old}}}
}{
\pi_{\theta_{\mathrm{old}}}(o_{i,t}\mid q,o_{i,<t})
}
=
\nabla_\theta
\log \pi_\theta(o_{i,t}\mid q,o_{i,<t})
\Big|_{\theta=\theta_{\mathrm{old}}}.
\]
Define the token-gradient vector
\[
v_{i,t}
:=
\nabla_\theta
\log \pi_\theta(o_{i,t}\mid q,o_{i,<t})
\Big|_{\theta=\theta_{\mathrm{old}}}.
\]
Then
\[
\nabla_\theta J_{\mathrm{DAPO}}(\theta)\Big|_{\theta=\theta_{\mathrm{old}}}
=
\frac{1}{N}
\sum_{i=1}^G \sum_{t=1}^{|o_i|}
\hat A_i\,v_{i,t}.
\]

For a first-order gradient ascent step with learning rate \(\eta>0\),
\[
\Delta\theta
=
\eta
\nabla_\theta J_{\mathrm{DAPO}}(\theta)\Big|_{\theta=\theta_{\mathrm{old}}}
=
\frac{\eta}{N}
\sum_{i=1}^G \sum_{t=1}^{|o_i|}
\hat A_i\,v_{i,t}.
\]
Thus, up to a positive proportionality constant,
\[
\Delta\theta
\propto
\sum_{i=1}^G \sum_{t=1}^{|o_i|}
\hat A_i\,v_{i,t}.
\]
Separating this sum by the sign of the sequence-level advantage gives the
positive-negative decomposition used in Section~\ref{sec:motive}:
\[
\Delta\theta
\propto
\sum_{i:\hat A_i>0}\sum_{t=1}^{|o_i|}
\hat A_i\,v_{i,t}
-
\sum_{i:\hat A_i<0}\sum_{t=1}^{|o_i|}
|\hat A_i|\,v_{i,t}.
\]

Finally, we derive the induced local change in next-token log-probabilities.
Let \(c\) denote an arbitrary generation context, and let \(x\) be a candidate
next token under this context. Define
\[
\Delta \log \pi(x\mid c)
:=
\log \pi_{\theta_{\mathrm{old}}+\Delta\theta}(x\mid c)
-
\log \pi_{\theta_{\mathrm{old}}}(x\mid c).
\]
A first-order Taylor expansion around \(\theta_{\mathrm{old}}\) gives
\[
\Delta \log \pi(x\mid c)
\approx
\left(
\nabla_\theta
\log \pi_\theta(x\mid c)
\Big|_{\theta=\theta_{\mathrm{old}}}
\right)^{\top}
\Delta\theta.
\]
Substituting the local DAPO update yields
\[
\Delta \log \pi(x\mid c)
\propto
\left(
\nabla_\theta
\log \pi_\theta(x\mid c)
\Big|_{\theta=\theta_{\mathrm{old}}}
\right)^{\top}
\sum_{i=1}^G \sum_{t=1}^{|o_i|}
\hat A_i\,v_{i,t}.
\]
Equivalently, using the positive-negative decomposition,
\[
\Delta \log \pi(x\mid c)
\propto
\left(
\nabla_\theta
\log \pi_\theta(x\mid c)
\Big|_{\theta=\theta_{\mathrm{old}}}
\right)^{\top}
\left(
\sum_{i:\hat A_i>0}\sum_{t=1}^{|o_i|}
\hat A_i\,v_{i,t}
-
\sum_{i:\hat A_i<0}\sum_{t=1}^{|o_i|}
|\hat A_i|\,v_{i,t}
\right).
\]
This is the first-order next-token log-probability change analyzed in Section~\ref{sec:motive}.

This expression describes the raw surrogate-gradient direction before optimizer
preconditioning, weight decay, or gradient clipping. In practical training with
Adam-type optimizers, the actual parameter step may apply an additional
preconditioner to this gradient. Our analysis focuses on the update direction
induced by the RLVR surrogate itself.

A degenerate case occurs when one advantage side is empty. Under group-normalized advantages, if there is no positive-advantage response in a group, then there is also no negative-advantage response: all advantages in the group are zero. Such a group contributes zero to the local policy-gradient update and is therefore skipped in the positive-negative decomposition. Equivalently, the sums over empty sides are treated as zero, and centroid quantities that require nonzero side mass are computed only for rollout groups with both positive and negative advantage sides.

\section{Mean Directions as Weighted Least-Squares Centroids}
\label{app:l2-centroid}

In this appendix, we show that the side-wise mean directions used in the
local discriminator view are exactly the weighted least-squares centroids
of the positive and negative sides.

Recall the local token-gradient vectors
\[
v_{i,t}
:=
\nabla_\theta
\log \pi_\theta(o_{i,t}\mid q,o_{i,<t})
\Big|_{\theta=\theta_{\mathrm{old}}},
\]
where \(i\) indexes responses in the rollout group and \(t\) indexes valid
response tokens. We split tokens according to the sign of the response-level
advantage \(\hat A_i\). Define the total positive and negative advantage masses as
\[
M_+
:=
\sum_{i:\hat A_i>0}\sum_{t=1}^{|o_i|}
\hat A_i,
\qquad
M_-
:=
\sum_{i:\hat A_i<0}\sum_{t=1}^{|o_i|}
|\hat A_i|.
\]
When \(M_+>0\) and \(M_->0\), the corresponding normalized side-wise mean
directions are
\[
\bar\mu_+
=
\frac{
\sum_{i:\hat A_i>0}\sum_{t=1}^{|o_i|}
\hat A_i v_{i,t}
}{
M_+
},
\qquad
\bar\mu_-
=
\frac{
\sum_{i:\hat A_i<0}\sum_{t=1}^{|o_i|}
|\hat A_i| v_{i,t}
}{
M_-
}.
\]
These are the positive- and negative-side reference directions used by the
local discriminator.

We now show that \(\bar\mu_+\) and \(\bar\mu_-\) are weighted least-squares
centroids. Consider first the positive side and define
\[
F_+(\mu)
:=
\sum_{i:\hat A_i>0}\sum_{t=1}^{|o_i|}
\hat A_i\,\|v_{i,t}-\mu\|_2^2.
\]
Differentiating with respect to \(\mu\) gives
\[
\nabla_\mu F_+(\mu)
=
2\sum_{i:\hat A_i>0}\sum_{t=1}^{|o_i|}
\hat A_i(\mu-v_{i,t})
=
2M_+\mu
-
2\sum_{i:\hat A_i>0}\sum_{t=1}^{|o_i|}
\hat A_i v_{i,t}.
\]
Setting the gradient to zero yields
\[
\mu
=
\frac{
\sum_{i:\hat A_i>0}\sum_{t=1}^{|o_i|}
\hat A_i v_{i,t}
}{
M_+
}
=
\bar\mu_+.
\]
Moreover,
\[
\nabla_\mu^2 F_+(\mu)
=
2M_+ I,
\]
which is positive definite whenever \(M_+>0\). Thus, \(\bar\mu_+\) is the
unique minimizer of \(F_+(\mu)\).

The negative side follows analogously. Define
\[
F_-(\mu)
:=
\sum_{i:\hat A_i<0}\sum_{t=1}^{|o_i|}
|\hat A_i|\,\|v_{i,t}-\mu\|_2^2.
\]
Then
\[
\nabla_\mu F_-(\mu)
=
2M_-\mu
-
2\sum_{i:\hat A_i<0}\sum_{t=1}^{|o_i|}
|\hat A_i| v_{i,t}.
\]
The first-order optimality condition gives
\[
\mu
=
\frac{
\sum_{i:\hat A_i<0}\sum_{t=1}^{|o_i|}
|\hat A_i| v_{i,t}
}{
M_-
}
=
\bar\mu_-.
\]
Since
\[
\nabla_\mu^2 F_-(\mu)
=
2M_- I
\succ 0
\]
whenever \(M_->0\), \(\bar\mu_-\) is also the unique minimizer.

Therefore, the side-wise mean directions \(\bar\mu_+\) and \(\bar\mu_-\)
are exactly the weighted least-squares centroids of the positive and negative
token-gradient vectors, respectively. If one side has zero total mass, the
corresponding centroid is undefined; such degenerate groups contribute no
positive-negative centroid contrast and are skipped in the centroid
construction.

\section{Representative RLVR Variants as Token-Weighted Centroid Estimators}
\label{app:centroid-view-rlvr}

In this appendix, we show that several representative RLVR variants can be
interpreted through the same local centroid view. The goal is not to claim that
these methods are identical, but to clarify how their local update rules modify
the effective token weights used to estimate the positive and negative
directions.

Consider a generic RLVR surrogate whose unclipped local update around the old
policy can be written as
\begin{equation}
\label{eq:generic-local-update}
\Delta\theta
\propto
\sum_{i=1}^{G}\sum_{t=1}^{|o_i|}
\rho_{i,t}\hat A_i v_{i,t},
\end{equation}
where
\[
v_{i,t}
=
\nabla_\theta
\log \pi_\theta(o_{i,t}\mid q,o_{i,<t})
\Big|_{\theta=\theta_{\mathrm{old}}},
\]
and \(\rho_{i,t}\ge 0\) denotes the effective token weight induced by the
objective. Global normalizers shared by all tokens are omitted when they do not
affect the update direction. 

Throughout this appendix, the effective token weights \(\rho_{i,t}\) are treated
as fixed stop-gradient quantities. Thus, the local update keeps only the gradient through the policy
ratio \(r_{i,t}(\theta)\). 

Separating positive- and negative-advantage responses gives
\begin{equation}
\label{eq:generic-centroid-decomp}
\Delta\theta
\propto
M_+(\rho)\mu_+(\rho)
-
M_-(\rho)\mu_-(\rho),
\end{equation}
where
\begin{equation}
\label{eq:generic-centroids}
\mu_+(\rho)
=
\frac{
\sum_{i:\hat A_i>0}\sum_{t=1}^{|o_i|}
\rho_{i,t}\hat A_i v_{i,t}
}{
\sum_{i:\hat A_i>0}\sum_{t=1}^{|o_i|}
\rho_{i,t}\hat A_i
},
\qquad
\mu_-(\rho)
=
\frac{
\sum_{i:\hat A_i<0}\sum_{t=1}^{|o_i|}
\rho_{i,t}|\hat A_i| v_{i,t}
}{
\sum_{i:\hat A_i<0}\sum_{t=1}^{|o_i|}
\rho_{i,t}|\hat A_i|
}.
\end{equation}
Here \(M_+(\rho)\) and \(M_-(\rho)\) are the corresponding total positive and
negative masses, i.e., the denominators of the two centroids above. Therefore,
under the local view, changing \(\rho_{i,t}\) changes both the side-wise
centroids and their relative masses.

\paragraph{GRPO.}
GRPO averages token losses within each response and then averages over
responses. Around the old policy, where clipping is locally inactive, its update
direction is
\begin{equation}
\label{eq:grpo-local-centroid}
\Delta\theta_{\mathrm{GRPO}}
\propto
\frac{1}{G}
\sum_{i=1}^{G}
\frac{\hat A_i}{|o_i|}
\sum_{t=1}^{|o_i|}
v_{i,t}.
\end{equation}
Thus, up to a global constant,
\[
\rho_{i,t}^{\mathrm{GRPO}}=\frac{1}{|o_i|}.
\]
Equivalently, GRPO first forms a response-level average token-gradient vector,
\[
\bar v_i
=
\frac{1}{|o_i|}
\sum_{t=1}^{|o_i|}
v_{i,t},
\]
and then aggregates these response-level vectors with group-relative advantages:
\[
\Delta\theta_{\mathrm{GRPO}}
\propto
\sum_{i=1}^{G}\hat A_i \bar v_i.
\]
Therefore, each response contributes one averaged direction. Longer responses
have smaller per-token weights, while shorter responses have larger per-token
weights. In the centroid view, GRPO estimates side-wise directions from
response-balanced token-gradient averages.

\paragraph{DAPO.}
DAPO replaces response-level averaging with token-level normalization over all
valid response tokens. Locally, its update direction is
\begin{equation}
\label{eq:dapo-local-centroid}
\Delta\theta_{\mathrm{DAPO}}
\propto
\sum_{i=1}^{G}
\sum_{t=1}^{|o_i|}
\hat A_i v_{i,t},
\end{equation}
where the global token-count normalizer is omitted. Hence
\[
\rho_{i,t}^{\mathrm{DAPO}}=1.
\]
Compared with GRPO, DAPO gives equal weight to every valid token. Consequently,
the total contribution of a response is proportional to its length. In the
centroid view, DAPO changes the side-wise estimator from a response-balanced
average to a token-balanced average.

\paragraph{DAPO with forking tokens.}
DAPO with forking tokens keeps only high-entropy tokens in the policy-gradient
loss. Let
\[
m_{i,t}^{\mathrm{FT}}
=
\mathbb{I}\!\left[H_{i,t}\ge \tau_\rho^{\mathcal B}\right]
\]
denote the high-entropy token mask in the current batch. The local update becomes
\begin{equation}
\label{eq:ft-local-centroid}
\Delta\theta_{\mathrm{FT}}
\propto
\sum_{i=1}^{G}
\sum_{t=1}^{|o_i|}
m_{i,t}^{\mathrm{FT}}\hat A_i v_{i,t}.
\end{equation}
Thus,
\[
\rho_{i,t}^{\mathrm{FT}}=m_{i,t}^{\mathrm{FT}}.
\]
This objective changes the support of the centroid estimator: low-entropy tokens
are removed, and the positive and negative centroids are estimated only from
high-entropy tokens. In this view, forking-token filtering emphasizes
high-entropy tokens as the main contributors to the side-wise directions.

\paragraph{FIPO.}
FIPO introduces a future-influence weight for each token. In its objective, the
standard advantage is multiplied by a Future-KL importance weight \(f_{i,t}\),
and this weighted advantage is used inside a DAPO-style token-level surrogate.
FIPO defines this weight from a discounted future probability-shift signal and
clips it to control variance.

Ignoring clipping locally, the update direction is
\begin{equation}
\label{eq:fipo-local-centroid}
\Delta\theta_{\mathrm{FIPO}}
\propto
\sum_{i=1}^{G}
\sum_{t=1}^{|o_i|}
f_{i,t}\hat A_i v_{i,t}.
\end{equation}
Therefore,
\[
\rho_{i,t}^{\mathrm{FIPO}}=f_{i,t}.
\]
In the centroid view, FIPO estimates the positive and negative centroids from
future-influence-weighted token-gradient vectors. Tokens with larger
future-influence weights receive larger centroid mass. This differs from entropy
filtering: FIPO does not simply select uncertain tokens, but weights tokens
according to a forward-looking estimate of their influence on later trajectory
behavior.

\paragraph{Summary.}
Under a local first-order view, several representative RLVR variants can be
interpreted as modifying the effective token weights \(\rho_{i,t}\) in an
advantage-weighted token-gradient aggregation. These weights determine the
side-wise centroids \(\mu_+(\rho)\) and \(\mu_-(\rho)\), as well as their total
masses. Existing methods adjust this estimator through length normalization,
entropy filtering, or future-influence weighting. DelTA differs by assigning
token weights according to whether a token gradient is more representative of
its own side than of the opposite side, thereby aligning the centroid estimator
with the discriminative structure of the local DAPO update.

\section{Last-layer Token-Gradient Proxy}
\label{app:proxy}

For scalable coefficient estimation, we use a layer-restricted token-gradient proxy based on the standard fixed-representation view of the output layer. Let the LM head produce logits \(z_t = W h_t\), where \(h_t \in \mathbb{R}^d\)
is the final hidden state at step \(t\), \(W \in \mathbb{R}^{|\mathcal V|\times d}\)
is the LM-head matrix, and \(p_t=\mathrm{softmax}(z_t)\). For the realized token
\(y_t\), the token log-probability is
\[
\log p_t(y_t)
=
W_{y_t}^{\top}h_t
-
\log\sum_j \exp(W_j^{\top}h_t).
\]
Differentiating with respect to the corresponding LM-head row gives
\[
\nabla_{W_{y_t}}\log p_t(y_t)
=
\bigl(1-p_t(y_t)\bigr)h_t.
\]
Thus, under a frozen-representation approximation, we use
\(\bigl(1-p_t(y_t)\bigr)h_t\) as a layer-restricted proxy for the token-gradient
vector.

We do not use this proxy as an exact reconstruction of the full-parameter gradient.  Instead, we use it as a scalable layer-restricted representation for estimating relative token coefficients. It is a pragmatic reduction that makes token-level analysis and reweighting feasible in large autoregressive language models. Prior work on gradient-based influence has similarly relied on first-order approximations and layer-restricted computations for scalability~\citep{pruthi2020estimating}, and in NLP it is common to restrict such analysis to the last layer when
full-parameter computations are impractical at scale~\citep{yeh2022first}.
This restriction should therefore be understood as an approximation, not an
exact reconstruction of the full gradient signal~\citep{yeh2022first}.

\paragraph{Proxy ablations.}
In addition to the default proxy above, we ablate the choice of layer-restricted
proxy used for DelTA coefficient estimation. These ablations do not change the
DelTA objective or the policy-gradient update; they only replace the
stop-gradient proxy vectors used to compute centroids, distances, and token
coefficients.

Let \(\mathcal K_t\) denote the top-\(K\) vocabulary indices under the current
logits, and let \(\widetilde p_t\) be the softmax distribution renormalized
within \(\mathcal K_t\). We consider a top-\(K\) hidden-gradient proxy
\[
\widehat v^{\mathrm{hid}}_t
=
W_{y_t}
-
\sum_{j\in\mathcal K_t}
\widetilde p_t(j) W_j,
\]
which is a top-\(K\) approximation to
\(\nabla_{h_t}\log p_t(y_t)\). Compared with the default output-row proxy
\(\bigl(1-p_t(y_t)\bigr)h_t\), this proxy explicitly incorporates the competing
high-probability tokens under the current policy.

We also include a random-coefficient baseline to test whether DelTA benefits
merely from perturbing token weights. In this baseline, \(\lambda_{i,t}\) is
sampled randomly from the same bounded range
\([\lambda_{\min},\lambda_{\max}]\), and then used with the same coefficient
normalization and weighted DAPO objective as DelTA. All other training settings
are kept unchanged.
\begin{table}[t]
\centering
\caption{
Proxy ablation study of DelTA. 
The best results are in \textbf{bold}.
}
\label{tab:proxy-ablation}
\small
\setlength{\tabcolsep}{4pt}
\renewcommand{\arraystretch}{1.06}

\begin{tabularx}{\linewidth}{@{}>{\raggedright\arraybackslash}p{0.34\linewidth}*{5}{>{\centering\arraybackslash}X}@{}}
\toprule
\textbf{Method}
& AIME25
& AIME26
& HMMT25
& HMMT26
& Avg. \\
\midrule

\rowcolor{cyan!10}
\bodystrut \textbf{Base DelTA}
& 26.46 & \textbf{28.12} & 18.54 & 20.27 & 23.27 \\

\midrule
\bodystrut Top-\(K\) hidden-gradient proxy
& \textbf{27.08} & 27.71 & \textbf{20.83} & \textbf{21.78} & \textbf{24.29} \\
\bodystrut Random \(\lambda\)
& 22.50 & 22.50 & 11.87 & 16.67 & 18.34 \\

\bottomrule
\end{tabularx}
\end{table}

Table~\ref{tab:proxy-ablation} shows that DelTA is robust to the choice of
last-layer proxy. The top-\(K\) hidden-gradient proxy achieves the best average
performance, improving over the default proxy from \(23.27\) to \(24.29\).
This suggests that the contrast-sensitive reweighting mechanism does not rely
on a single proxy choice, and that incorporating local competition among top
vocabulary candidates can provide a stronger signal for coefficient estimation.

In contrast, the random-\(\lambda\) baseline substantially underperforms both
DelTA variants, dropping to an average score of \(18.34\). This confirms that
the benefit of DelTA is not due to arbitrary token reweighting or stochastic
perturbation of the loss. Rather, the coefficients need to preserve meaningful
discriminative structure: tokens should receive larger weights when their
gradients are more representative of their own advantage side than of the
opposite side.

Therefore, the gains reported in our main experiments are obtained with a
conservative and computationally minimal proxy. The stronger performance of the
top-\(K\) hidden-gradient variant further suggests that DelTA is not tied to this
specific proxy choice and may benefit from more informative gradient proxies. We keep the output-row proxy as the default in the main experiments because it is the minimal and most direct layer-restricted instantiation of the token-gradient view; the stronger top-\(K\) proxy shows that DelTA can further benefit from richer proxy choices.

\section{Derivation of the DelTA Soft Assignment Score}
\label{app:DelTA-weight-derivation}

In this appendix, we derive the closed-form solution of the soft assignment
score used in DelTA. We follow the notation in Section~\ref{sec:DelTA}. The
centroids \(\mu_+\) and \(\mu_-\) are fixed in this derivation, and
\(\alpha_{i,t}\in[0,1]\) denotes the raw assignment score before the final
remapping and normalization step.

\paragraph{Side-specific squared-distance margin.}
For a token from a positive-advantage response, DelTA assigns a larger score
when its token-gradient vector is closer to the positive centroid than to the
negative centroid. We define the positive-side squared-distance margin as
\[
\Delta_{i,t}^{+}
:=
\|v_{i,t}-\mu_-\|_2^2
-
\|v_{i,t}-\mu_+\|_2^2,
\qquad
\hat A_i>0.
\]
Thus, \(\Delta_{i,t}^{+}>0\) means that \(v_{i,t}\) is closer to \(\mu_+\) than
to \(\mu_-\). For a token from a negative-advantage response, the two centroids
are swapped, and we define
\[
\Delta_{i,t}^{-}
:=
\|v_{i,t}-\mu_+\|_2^2
-
\|v_{i,t}-\mu_-\|_2^2,
\qquad
\hat A_i<0.
\]
Hence, \(\Delta_{i,t}^{-}>0\) means that \(v_{i,t}\) is closer to \(\mu_-\) than
to \(\mu_+\).

\paragraph{Positive-side assignment.}
For a token from a positive-advantage response,
\Eq{eq:DelTA-assignment-objective} can be written compactly as
\[
\alpha_{i,t}
=
\arg\max_{\alpha\in[0,1]}
\alpha \Delta_{i,t}^{+}
+
\gamma_+ h(\alpha),
\]
where
\[
h(\alpha)
=
-\alpha\log\alpha-(1-\alpha)\log(1-\alpha)
\]
is the binary entropy, with the standard convention \(0\log 0=0\). Since the
centroids and temperature are fixed, this is a one-dimensional optimization
problem. To simplify notation, write
\(\Delta=\Delta_{i,t}^{+}\) and \(\gamma=\gamma_+\). We solve
\[
\max_{\alpha\in[0,1]}
f(\alpha)
:=
\alpha\Delta+\gamma h(\alpha).
\]

For \(\alpha\in(0,1)\), the derivative is
\[
f'(\alpha)
=
\Delta+\gamma h'(\alpha)
=
\Delta+\gamma\log\frac{1-\alpha}{\alpha}.
\]
Setting \(f'(\alpha)=0\) gives
\[
\log\frac{1-\alpha}{\alpha}
=
-\frac{\Delta}{\gamma},
\]
or equivalently
\[
\frac{\alpha}{1-\alpha}
=
\exp\!\left(\frac{\Delta}{\gamma}\right).
\]
Solving for \(\alpha\) yields
\[
\alpha
=
\frac{1}{1+\exp(-\Delta/\gamma)}
=
\sigma\!\left(\frac{\Delta}{\gamma}\right),
\]
where \(\sigma(z)=1/(1+\exp(-z))\).

The solution is the unique maximizer. Indeed,
\[
f''(\alpha)
=
\gamma h''(\alpha)
=
-\gamma
\left(
\frac{1}{\alpha}
+
\frac{1}{1-\alpha}
\right)
=
-\frac{\gamma}{\alpha(1-\alpha)}
<0
\]
for all \(\alpha\in(0,1)\), since \(\gamma>0\). Therefore, \(f\) is strictly
concave on \((0,1)\), and the stationary point above is the unique optimum.
Substituting back \(\Delta=\Delta_{i,t}^{+}\) and \(\gamma=\gamma_+\), we obtain
\[
\alpha_{i,t}
=
\sigma\!\left(
\dfrac{
\|v_{i,t}-\mu_-\|_2^2
-
\|v_{i,t}-\mu_+\|_2^2
}{
\gamma_+
}
\right),
\qquad
\hat A_i>0.
\]

\paragraph{Negative-side assignment.}
For tokens from negative-advantage responses, DelTA uses the symmetric objective
obtained by swapping the positive and negative centroids:
\[
\alpha_{i,t}
=
\arg\max_{\alpha\in[0,1]}
\alpha \Delta_{i,t}^{-}
+
\gamma_- h(\alpha),
\qquad
\hat A_i<0.
\]
Repeating the same one-dimensional derivation with
\(\Delta=\Delta_{i,t}^{-}\) and \(\gamma=\gamma_-\) gives
\[
\alpha_{i,t}
=
\sigma\!\left(
\dfrac{
\|v_{i,t}-\mu_+\|_2^2
-
\|v_{i,t}-\mu_-\|_2^2
}{
\gamma_-
}
\right),
\qquad
\hat A_i<0.
\]

Combining the two sides, the DelTA soft assignment score is
\[
\alpha_{i,t}
=
\begin{cases}
\sigma\!\left(
\dfrac{
\|v_{i,t}-\mu_-\|_2^2
-
\|v_{i,t}-\mu_+\|_2^2
}{
\gamma_+
}
\right),
& \hat A_i>0,\\[1.2ex]
\sigma\!\left(
\dfrac{
\|v_{i,t}-\mu_+\|_2^2
-
\|v_{i,t}-\mu_-\|_2^2
}{
\gamma_-
}
\right),
& \hat A_i<0,
\end{cases}
\]
which matches \Eq{eq:DelTA-alpha}.

The score \(\alpha_{i,t}\) is a raw stop-gradient assignment score. As described
in Appendix~\ref{app:DelTA-iteration}, DelTA subsequently remaps these scores
to a bounded coefficient range and normalizes them before using them as loss
coefficients in the weighted DAPO surrogate.

\paragraph{Connection to the inner-product discriminator.}
For fixed centroids \(\mu_+\) and \(\mu_-\), the positive-side squared-distance margin used by DelTA can be rewritten as
\[
\|v-\mu_-\|_2^2-\|v-\mu_+\|_2^2
=
2v^\top(\mu_+-\mu_-)
+
\|\mu_-\|_2^2-\|\mu_+\|_2^2 .
\]
Thus, up to a centroid-dependent offset, the distance margin scores \(v\) by its alignment with the centroid contrast direction \(\mu_+-\mu_-\). 
The negative-side margin is obtained symmetrically by swapping \(\mu_+\) and \(\mu_-\). 
This shows that the squared-distance comparison used for coefficient estimation is consistent with the positive-negative inner-product discriminator view in Section~\ref{sec:motive}, while matching the weighted least-squares geometry of the side-wise centroids.

\section{DelTA Implementation Details}
\label{app:DelTA-iteration}

In this appendix, we describe how DelTA is computed in practice on each rollout batch.
The raw assignment scores depend on the side-wise centroids, while the centroids are themselves estimated using these scores.
We therefore use a small fixed number of lagged alternating refinement steps rather than solving the coupled problem to convergence.

Let \(\{v_{i,t}\}\) denote the token-gradient vectors in the rollout batch, and let \(\{\hat A_i\}\) denote the corresponding sequence-level advantages.
Throughout this appendix, \(i\in\{1,\dots,G\}\) indexes sampled responses and \(t\in\{1,\dots,|o_i|\}\) indexes valid response tokens.
We use \(\alpha_{i,t}\in[0,1]\) for raw assignment scores, and reserve \(\lambda_{i,t}\) for the final training coefficient used in the weighted DAPO surrogate.
All distances below are standard squared Euclidean distances.

\paragraph{Initialization of side-wise centroids.}
We initialize the positive and negative centroids from the original advantage-weighted token-gradient means.
Define
\[
M_+
=
\sum_{i:\hat A_i>0}\sum_{t=1}^{|o_i|}
\hat A_i,
\qquad
M_-
=
\sum_{i:\hat A_i<0}\sum_{t=1}^{|o_i|}
|\hat A_i|.
\]
The initial centroids are
\[
\mu_+^{(0)}
=
\frac{
\sum_{i:\hat A_i>0}\sum_{t=1}^{|o_i|}
\hat A_i\,v_{i,t}
}{
M_+
},
\qquad
\mu_-^{(0)}
=
\frac{
\sum_{i:\hat A_i<0}\sum_{t=1}^{|o_i|}
|\hat A_i|\,v_{i,t}
}{
M_-
}.
\]
In implementation, the denominators are clamped by a small numerical constant \(\varepsilon>0\) for stability.

\paragraph{Squared-distance differences and lagged adaptive temperatures.}
For centroids \(\mu_+^{(k)}\) and \(\mu_-^{(k)}\), define the positive-side and negative-side squared-distance differences as
\[
\Delta_{i,t}^{+,(k)}
=
\|v_{i,t}-\mu_-^{(k)}\|_2^2
-
\|v_{i,t}-\mu_+^{(k)}\|_2^2,
\qquad \hat A_i>0,
\]
and
\[
\Delta_{i,t}^{-,(k)}
=
\|v_{i,t}-\mu_+^{(k)}\|_2^2
-
\|v_{i,t}-\mu_-^{(k)}\|_2^2,
\qquad \hat A_i<0.
\]
A larger value means that the token-gradient vector is closer to its own-side centroid than to the opposite-side centroid.

The side-specific temperatures are set from the empirical standard deviations of these squared-distance differences.
For the initial centroids, we compute
\[
\gamma_+^{(0)}
=
\sqrt{
\max\!\left(
\operatorname{Var}\!\left(
\{\Delta_{i,t}^{+,(0)}:\hat A_i>0,\;1\le t\le |o_i|\}
\right),
\varepsilon_\gamma
\right)
},
\]
\[
\gamma_-^{(0)}
=
\sqrt{
\max\!\left(
\operatorname{Var}\!\left(
\{\Delta_{i,t}^{-,(0)}:\hat A_i<0,\;1\le t\le |o_i|\}
\right),
\varepsilon_\gamma
\right)
},
\]
where \(\varepsilon_\gamma>0\) is a small numerical constant.

The superscript of \(\gamma_\pm^{(k)}\) denotes the cached temperature used when computing \(\alpha_{i,t}^{(k)}\).
In implementation, the temperatures are updated in a lagged manner.
During the pass that computes \(\alpha_{i,t}^{(k)}\), we also accumulate the empirical variances of \(\Delta_{i,t}^{+,(k)}\) and \(\Delta_{i,t}^{-,(k)}\), and store the resulting temperatures as \(\gamma_+^{(k+1)}\) and \(\gamma_-^{(k+1)}\) for the next score-computation step.
Thus, for \(k>0\), \(\gamma_\pm^{(k)}\) is the cached temperature produced by the previous margin-statistics pass.
This avoids an additional proxy forward pass solely for temperature estimation.

\paragraph{Alternating refinement.}
Starting from \(\mu_+^{(0)}\), \(\mu_-^{(0)}\), \(\gamma_+^{(0)}\), and \(\gamma_-^{(0)}\), DelTA runs \(K\) stop-gradient refinement iterations.
At iteration \(k=0,\dots,K-1\), the raw assignment scores are
\[
\alpha_{i,t}^{(k)}
=
\begin{cases}
\sigma\!\left(
\Delta_{i,t}^{+,(k)}/\gamma_+^{(k)}
\right),
& \hat A_i>0,\\[1.0ex]
\sigma\!\left(
\Delta_{i,t}^{-,(k)}/\gamma_-^{(k)}
\right),
& \hat A_i<0,
\end{cases}
\]
where \(\sigma(\cdot)\) is the sigmoid function.
Equivalently, this is the closed-form solution of the entropy-regularized assignment objective in Section~\ref{sec:DelTA}.

The centroids are then updated as score-weighted within-side averages:
\[
\mu_+^{(k+1)}
=
\frac{
\sum_{i:\hat A_i>0}\sum_{t=1}^{|o_i|}
\hat A_i\,\alpha_{i,t}^{(k)}v_{i,t}
}{
\sum_{i:\hat A_i>0}\sum_{t=1}^{|o_i|}
\hat A_i\,\alpha_{i,t}^{(k)}
},
\]
\[
\mu_-^{(k+1)}
=
\frac{
\sum_{i:\hat A_i<0}\sum_{t=1}^{|o_i|}
|\hat A_i|\,\alpha_{i,t}^{(k)}v_{i,t}
}{
\sum_{i:\hat A_i<0}\sum_{t=1}^{|o_i|}
|\hat A_i|\,\alpha_{i,t}^{(k)}
}.
\]
Again, denominators are clamped by a small numerical constant for stability.
The same pass also accumulates the statistics of \(\Delta_{i,t}^{+,(k)}\) and \(\Delta_{i,t}^{-,(k)}\), producing \(\gamma_+^{(k+1)}\) and \(\gamma_-^{(k+1)}\) for the next refinement iteration.

\paragraph{Final coefficient computation.}
After \(K\) refinement iterations, we obtain refined centroids \(\mu_+^{(K)}\) and \(\mu_-^{(K)}\).
A final pass over the rollout batch recomputes the squared-distance differences using these refined centroids:
\[
\Delta_{i,t}^{+,(\star)}
=
\|v_{i,t}-\mu_-^{(K)}\|_2^2
-
\|v_{i,t}-\mu_+^{(K)}\|_2^2,
\qquad \hat A_i>0,
\]
\[
\Delta_{i,t}^{-,(\star)}
=
\|v_{i,t}-\mu_+^{(K)}\|_2^2
-
\|v_{i,t}-\mu_-^{(K)}\|_2^2,
\qquad \hat A_i<0.
\]
The final raw scores are computed using the latest cached temperatures:
\[
\alpha_{i,t}^{\star}
=
\begin{cases}
\sigma\!\left(
\Delta_{i,t}^{+,(\star)}/\gamma_+^{(K)}
\right),
& \hat A_i>0,\\[1.0ex]
\sigma\!\left(
\Delta_{i,t}^{-,(\star)}/\gamma_-^{(K)}
\right),
& \hat A_i<0.
\end{cases}
\]
When \(K=0\), this reduces to using the initial centroids \(\mu_+^{(0)}\), \(\mu_-^{(0)}\) and initial temperatures \(\gamma_+^{(0)}\), \(\gamma_-^{(0)}\).

The raw scores are then mapped to bounded loss coefficients:
\[
\lambda_{i,t}
=
\lambda_{\min}
+
(\lambda_{\max}-\lambda_{\min})\alpha_{i,t}^{\star},
\qquad \hat A_i\ne 0.
\]
For zero-advantage responses, which do not belong to either side, we assign
\[
\lambda_{i,t}=\lambda_{\min}.
\]

\paragraph{Self-normalized implementation with the standard DAPO token average.}
The self-normalized DelTA objective in \Eq{eq:DelTA-weighted-objective} can be implemented while keeping the standard DAPO token-count normalizer.
Let
\[
N
=
\sum_{i=1}^{G}|o_i|,
\qquad
Z
=
\sum_{i=1}^{G}\sum_{t=1}^{|o_i|}\lambda_{i,t}.
\]
We define the implementation-facing coefficient
\[
\bar\lambda_{i,t}
=
\lambda_{i,t}\frac{N}{Z}.
\]
By construction,
\[
\frac{1}{N}
\sum_{i=1}^{G}\sum_{t=1}^{|o_i|}
\bar\lambda_{i,t}
=
1.
\]
Therefore, for the clipped DAPO token loss
\[
\ell_{i,t}(\theta)
=
\min\!\Big(
r_{i,t}(\theta)\hat A_i,\;
\mathrm{clip}\!\big(
r_{i,t}(\theta),
1-\epsilon_{\mathrm{low}},
1+\epsilon_{\mathrm{high}}
\big)\hat A_i
\Big),
\]
we have
\[
\frac{1}{N}
\sum_{i=1}^{G}\sum_{t=1}^{|o_i|}
\bar\lambda_{i,t}\ell_{i,t}(\theta)
=
\frac{1}{Z}
\sum_{i=1}^{G}\sum_{t=1}^{|o_i|}
\lambda_{i,t}\ell_{i,t}(\theta).
\]
Thus, using \(\bar\lambda_{i,t}\) under the standard DAPO token average is exactly equivalent to the self-normalized DelTA objective.
The normalization changes only the global scale of the coefficients and preserves their relative token reweighting.

\paragraph{Use in training.}
The coefficients \(\lambda_{i,t}\) and \(\bar\lambda_{i,t}\) are treated as stop-gradient quantities.
They are computed once from the rollout batch, held fixed across repeated optimization passes over that batch, and recomputed only when new trajectories are sampled.
In the actual loss computation, \(\bar\lambda_{i,t}\) is used as the per-token multiplier, and the rest of the DAPO training pipeline remains unchanged.

\paragraph{Summary.}
For each rollout batch, DelTA proceeds as follows:
\begin{enumerate}
    \item Compute layer-restricted token-gradient proxies \(v_{i,t}\) for all valid rollout tokens.
    \item Initialize \(\mu_+^{(0)}\) and \(\mu_-^{(0)}\) from the original advantage-weighted side-wise centroids.
    \item Compute initial temperatures \(\gamma_+^{(0)}\) and \(\gamma_-^{(0)}\) from the squared-distance differences under the initial centroids.
    \item For \(k=0,\dots,K-1\):
    \begin{enumerate}
        \item compute raw assignments \(\alpha_{i,t}^{(k)}\) using \(\mu_+^{(k)}\), \(\mu_-^{(k)}\), and the cached temperatures \(\gamma_+^{(k)}\), \(\gamma_-^{(k)}\);
        \item update \(\mu_+^{(k+1)}\) and \(\mu_-^{(k+1)}\) as score-weighted within-side centroids;
        \item store \(\gamma_+^{(k+1)}\) and \(\gamma_-^{(k+1)}\) from the same squared-distance-difference statistics for the next assignment computation.
    \end{enumerate}
    \item Compute final raw assignments \(\alpha_{i,t}^{\star}\) using the refined centroids \(\mu_+^{(K)}\), \(\mu_-^{(K)}\), and latest cached temperatures \(\gamma_+^{(K)}\), \(\gamma_-^{(K)}\).
    \item Map \(\alpha_{i,t}^{\star}\) to bounded coefficients \(\lambda_{i,t}\), normalize them into \(\bar\lambda_{i,t}\), and use \(\bar\lambda_{i,t}\) inside the standard DAPO token average.
\end{enumerate}

\section{Detailed Settings}
\label{app:hyp}
The training settings are presented in Table~\ref{tab:RL-hyper-parameters}. We train the Qwen3-8B-Base model for 220 steps and the Qwen3-14B-Base model for 300 steps. For checkpoint selection, we evaluate checkpoints during training and select the checkpoint with the highest AIME25 avg@8 score. This selection rule is fixed in advance and applied uniformly to all methods and backbone sizes. Thus, our main comparison reflects the best checkpoint performance achieved by each method under the same training budget and the same model-selection protocol. 

All existing assets used in this work are publicly available research assets. We cite their original sources and follow their corresponding licenses and terms of use.

All experiments were conducted on \(8\times\) NVIDIA B200 GPUs. 
For the main DelTA training experiments, the wall-clock time per training step was approximately 5 minutes at the beginning of Qwen3-8B-Base training and increased to approximately 10 minutes as response lengths grew.
For Qwen3-14B-Base, the wall-clock time per training step was approximately 8 minutes at the beginning of training and increased to approximately 18 minutes as response lengths grew.

We use a simple binary verifiable reward: a response receives reward \(1\) if its final answer is correct, and \(0\) otherwise. Answer correctness is determined by \texttt{math-verify}\footnote{\url{https://github.com/huggingface/Math-Verify}}. Evaluation hyperparameters are presented in Table~\ref{tab:eval-hyperparameters}.
\begin{table}[t]  
\begin{center}
\begin{minipage}[t]{0.48\linewidth}
\caption{Training settings.}
\label{tab:RL-hyper-parameters}
\centering
\begin{tabular}{ll}
\toprule
Hyper-parameter & Value \\
\midrule
Train Batch Size & 128  \\
Micro Batch Size & 32 \\
Rollout $n$ & 16 \\
Maximum Prompt Length & 2048 \\
Maximum Response Length & 20,480 \\
Clip Ratio Low & 0.2 \\
Clip Ratio High & 0.28 \\
Rollout Engine & SGLang \\
Temperature & 1.0 \\
Top-p & 1.0 \\
LR & $1\times 10^{-6}$ \\
KL Coefficient & 0.0 \\
\bottomrule
\end{tabular}
\end{minipage}  
\hfil
\begin{minipage}[t]{0.48\linewidth}

\caption{Evaluation hyper-parameters.}
\label{tab:eval-hyperparameters}
\centering
\begin{tabular}{ll}
\toprule
Hyper-parameter & Value \\
\midrule
Max Length & 30,000  \\
Temperature & 1.0 \\
Top-p & 0.7 \\
Inference Engine & SGLang \\
\bottomrule
\end{tabular}
\end{minipage} 
\end{center}
\end{table}

\section{Baseline Details}
\label{app:baseline-details}

In this section, we briefly introduce the baseline methods used in our experiments.

\paragraph{DAPO.}
DAPO~\citep{yu2025dapo} extends GRPO by using asymmetric clipping, and token-level loss normalization to improve long-CoT RL training. We use DAPO as our main strong baseline.

\paragraph{DAPO w/ Forking Tokens.}
DAPO w/ Forking Tokens~\citep{wang2025beyond} (DAPO w/ FT for short) is a token-filtered variant of DAPO that keeps only the top \(20\%\) high-entropy tokens, referred to as forking tokens, in the policy-gradient loss. 
The remaining low-entropy tokens are masked out, based on the observation that high-entropy tokens often correspond to critical reasoning forks.

\paragraph{SAPO.}
SAPO~\citep{gao2025soft} replaces hard clipping with a smooth, temperature-controlled soft gate to attenuate off-policy updates. 
It further uses different temperatures for positive- and negative-advantage tokens to improve training stability.

\paragraph{FIPO.}
FIPO~\citep{ma2026fipo} reweights token-level advantages using discounted Future-KL, which estimates how much each token influences the subsequent trajectory. 
The resulting influence weights are clipped to control variance during policy optimization.

The method-specific hyperparameters of SAPO and FIPO are provided in Table~\ref{tab: SAPO hyper-parameters} and Table~\ref{tab: FIPO hyper-parameters}, respectively.

\begin{table}[t]  
\begin{center}
\begin{minipage}[t]{0.48\linewidth}
\caption{Unique training hyper-parameters of SAPO.}
\label{tab: SAPO hyper-parameters}
\centering
\begin{tabular}{ll}
\toprule
Hyper-parameter & Value \\
\midrule
Gae Gamma & 1.0 \\
Gae Lam & 0.95 \\
Tau Pos & 1.0 \\
Tau Neg & 1.05 \\
\bottomrule
\end{tabular}
\end{minipage}  
\hfil
\begin{minipage}[t]{0.48\linewidth}
\caption{Unique training hyper-parameters of FIPO.}
\label{tab: FIPO hyper-parameters}
\centering
\begin{tabular}{ll}
\toprule
Hyper-parameter & Value \\
\midrule
Decay Rate & 32.0  \\
Chunk Size & 128 \\
Future KL Start & include current \\
Future KL Window & -1 \\
Future KL Average & False \\
Future KL Clip Ratio & 0.2 \\
Future KL Clip High Only & True \\
Safety Thresh & 10.0 \\
\bottomrule
\end{tabular}
\end{minipage} 
\end{center}
\end{table}

\section{Significance Test Details}
\label{app:sig}

We describe the statistical significance tests used in our main comparison.
Due to the high cost of RLVR training, we do not repeat full training runs with multiple random seeds.
Instead, we perform an evaluation-run-level significance test that captures stochasticity from repeated generation.

For each method, we repeat the full evaluation \(S=16\) times.
Each repetition runs the model over the evaluation suite and produces one question-count-weighted aggregate score.
Since different methods are evaluated with independently sampled runs, we treat the 16 scores from DelTA and the 16 scores from the baseline as two independent samples.

We use a one-sided Mann--Whitney \(U\) test as the primary non-parametric test, with the pre-specified alternative hypothesis that DelTA outperforms the baseline. We regard the improvement as statistically significant when the one-sided \(p\)-value is below \(0.05\).

For the main significance claim, we compare DelTA with the strongest same-scale baseline for each backbone size, namely SAPO for the 8B backbone and FIPO for the 14B backbone.
Under this evaluation-run-level testing protocol, DelTA significantly outperforms the strongest same-scale baseline at both model scales.

\section{Supplementary experiments}

\subsection{Computational Overhead}
\label{app:expense}

In this subsection, we discuss the computational overhead introduced by DelTA.

The DelTA proxy is computed from final-layer hidden states, which are already
produced when evaluating old log-probabilities. Therefore, if all hidden states
for the rollout batch could be cached, the centroid refinement itself would not
require additional actor forward passes: the same fixed token-gradient proxies
could be reused across refinement iterations.

In practice, caching all hidden states for long-response RLVR rollouts is memory
intensive. We therefore recompute the proxy whenever it is needed. Relative to
the standard old-log-probability computation, the additional actor forward passes
for DelTA with \(K\) refinement iterations are:

\begin{enumerate}
    \item one pass to estimate the initial temperature scale \(\gamma\);
    \item one pass for each refinement iteration \(k=1,\dots,K\), which computes
    token scores and updates the centroids;
    \item one final pass to compute the final token coefficients used in the
    weighted DAPO objective.
\end{enumerate}

Thus, DelTA requires \(K+2\) additional actor forward passes for coefficient
estimation. In our main experiments, we use \(K=1\), which already gives strong
performance. These additional passes are only used to compute stop-gradient
coefficients; the resulting weighted DAPO objective is optimized in the same way
as the baseline objective.

To empirically measure the overhead, we compare the execution time of the first
training step between DelTA and DAPO. We focus on the first step because response
lengths can diverge during training, as shown in Figure~\ref{fig:training_dynamics}.
Since rollout generation dominates wall-clock time in long-response RLVR, later-step
end-to-end comparisons would be confounded by differences in generated response
length rather than isolating the overhead of DelTA's coefficient computation.

On 8 NVIDIA B200 GPUs, the first step of DelTA takes 37 seconds longer than
DAPO. Since rollout generation dominates long-response RLVR, this corresponds to approximately \(10.2\%\) of the total first-step time
of DelTA. These results indicate that, when isolating the optimization phase
from length-induced rollout variation, the overhead of DelTA's centroid
refinement is modest for RLVR workloads.

\subsection{Other Model Architectures}
\label{app:other-architectures}

To further examine the generality of DelTA beyond the Qwen3 backbones, we also conduct experiments on Olmo3-7B-Base.
We train DelTA and the DAPO baseline with the same hyperparameter settings as in the main experiments, without architecture-specific tuning.
This setting allows us to test whether the proposed token-level credit assignment remains effective on a different model family.

\begin{table*}[t]
\centering
\caption{Main results on seven mathematical reasoning benchmarks for Olmo3-7B-Base. The best results are in \textbf{bold}.}
\label{tab:results_olmo}
\small
\setlength{\tabcolsep}{2pt}
\renewcommand{\arraystretch}{1.06}

\begin{tabularx}{\textwidth}{@{}>{\raggedright\arraybackslash}p{0.16\textwidth}*{8}{>{\centering\arraybackslash}X}@{}}
\toprule
\textbf{Method}
& AIME24
& AIME25
& AIME26
& \makecell[c]{HMMT25\\(Feb.)}
& \makecell[c]{HMMT25\\(Nov.)}
& \makecell[c]{HMMT26\\(Feb.)}
& Brumo25
& Avg. \\
\midrule

\multicolumn{9}{@{}l@{}}{\emph{\textbf{Olmo3-7B-Base}}} \\
\bodystrut DAPO        & 25.62 & 21.88 & 22.08 & 14.58 & 8.75 & 14.02 & 26.67 & 19.01 \\

\rowcolor{cyan!10}
\bodystrut \textbf{DelTA} & \textbf{30.83} & \textbf{24.79} & \textbf{27.08} & \textbf{16.67}
              & \textbf{13.96} & \textbf{16.29} & \textbf{30.63} & \textbf{22.80} \\

\bottomrule
\end{tabularx}
\end{table*}

As shown in Table~\ref{tab:results_olmo}, DelTA consistently outperforms DAPO on all seven mathematical reasoning benchmarks.
The average score improves from \(19.01\) to \(22.80\), corresponding to a gain of \(3.79\) points.
The improvement is particularly large on AIME24, AIME26, HMMT25-Nov, and Brumo25, while DelTA also maintains clear gains on the remaining benchmarks.
These results suggest that DelTA is not tied to a specific Qwen3 backbone, and that discriminative token-level reweighting can provide consistent benefits across different base model families.

\subsection{Code Generation}
\label{app:code-gen}

\begin{wraptable}{r}{0.48\textwidth}
\vspace{-10pt}
\centering
\caption{Code generation results. DelTA consistently improves performance across all benchmarks.}
\label{tab:results_on_Code}
\scriptsize
\setlength{\tabcolsep}{3pt}
\renewcommand{\arraystretch}{0.85}

\begin{tabularx}{\linewidth}{
@{}>{\small\raggedright\arraybackslash}l
*{4}{>{\small\centering\arraybackslash}X}
@{}}
\toprule
{\scriptsize\textbf{Method}}
& {\scriptsize\textbf{HumanEval+}}
& {\scriptsize\textbf{MBPP+}}
& {\scriptsize\textbf{LCB}}
& {\scriptsize\textbf{Avg.}} \\
\midrule

DAPO & 83.0 & 72.1 & 33.4 & 47.7 \\
\rowcolor{cyan!10}
\textbf{DelTA} & \textbf{84.6} & \textbf{73.2} & \textbf{35.6} & \textbf{49.5} \\
\bottomrule
\end{tabularx}
\vspace{-10pt}
\end{wraptable}
To further validate the effectiveness of DelTA beyond mathematical reasoning, we conduct experiments on code generation tasks. We use Eurus2-RL-Code~\citep{cui2025process} as the training dataset and adopt DAPO as the baseline. All models are trained for two epochs under the same training recipe. We evaluate the trained models on HumanEval+, MBPP+~\citep{liu2023your}, and LiveCodeBench~\citep{jain2024livecodebench}. For each problem, we sample 5 rollouts and report the average accuracy. We also report a weighted average across benchmarks, where each benchmark is weighted by its number of evaluation problems.

As shown in Table~\ref{tab:results_on_Code}, DelTA consistently improves over DAPO on all three code generation benchmarks, increasing the weighted average score from 47.7 to 49.5. These results suggest that the benefit of DelTA is not limited to mathematical reasoning, but also transfers to code generation tasks, where effective token-level credit assignment remains important under sequence-level supervision.

\subsection{Q4: Is DelTA sensitive to its hyperparameters?}
\label{app:hyp-sensitivity}

\begin{table}[t]
\centering
\caption{
Hyperparameter sensitivity study of DelTA. 
The best results are in \textbf{bold}, and the second-best results are \underline{underlined}.
}
\label{tab:sens}
\small
\setlength{\tabcolsep}{4pt}
\renewcommand{\arraystretch}{1.06}

\begin{tabularx}{\linewidth}{@{}>{\raggedright\arraybackslash}p{0.34\linewidth}*{5}{>{\centering\arraybackslash}X}@{}}
\toprule
\textbf{Method}
& AIME25
& AIME26
& HMMT25
& HMMT26
& Avg. \\
\midrule

\rowcolor{cyan!10}
\bodystrut \textbf{Base DelTA}
& \textbf{26.46} & \underline{28.12} & \underline{18.54} & \textbf{20.27} & \textbf{23.27} \\

\midrule
\bodystrut \(\lambda_{\min} = 0.5\)
& 25.42 & \textbf{28.96} & \textbf{19.17} & 19.70 & \underline{23.22} \\
\bodystrut \(\lambda_{\max} = 1.5\)
& \underline{26.46} & 26.25 & 18.75 & 19.89 & 22.77 \\
\bodystrut \(\lambda_{\min} = 0.5, \lambda_{\max} = 1.5\)
& 25.62 & 27.92 & 18.96 & \underline{20.08} & 23.07 \\
\bodystrut \(K=2\)
& 25.00 & 27.08 & 18.33 & 18.56 & 22.15 \\
\bodystrut \(K=3\)
& 26.04 & 26.67 & 18.12 & 18.37 & 22.20 \\

\bottomrule
\end{tabularx}
\end{table}
We further study the sensitivity of DelTA to its method-specific hyperparameters.
We focus on two factors: the coefficient range \([\lambda_{\min}, \lambda_{\max}]\), which controls the strength of token reweighting, and the number of centroid refinement iterations \(K\), which controls how many times the side-wise centroids are updated before computing the final coefficients.
For the coefficient range, we evaluate several configurations, including  \([0.5,1.2]\), \([0.8,1.5]\), and \([0.5,1.5]\).
For the refinement step, we vary \(K\) over different values, including  \(K=2\) and \(K=3\).

As shown in Table~\ref{tab:sens}, DelTA is relatively robust to the coefficient range: changing \(\lambda_{\min}\) or \(\lambda_{\max}\) only leads to small average-score variations, and \(\lambda_{\min}=0.5\) achieves a comparable average score to the base setting.
In contrast, increasing the refinement depth to \(K=2\) or \(K=3\) consistently reduces performance, suggesting that a single refinement step is sufficient and that excessive refinement may overfit the batch-level token-gradient geometry.
Overall, the base configuration achieves the best average performance, indicating that \([\lambda_{\min},\lambda_{\max}]=[0.8,1.2]\) with \(K=1\) provides a stable trade-off between effective token reweighting and optimization robustness.

\subsection{Q5: Does DelTA generalize to out-of-domain evaluation?}
\label{app:ood}
\begin{wraptable}{r}{0.42\textwidth}
\centering
\caption{
OOD evaluation on GPQA-D and MMLU-Pro.
The best results are in \textbf{bold}.
}
\label{tab:results_on_ood}
\scriptsize
\setlength{\tabcolsep}{3.5pt}
\renewcommand{\arraystretch}{0.88}

\begin{tabularx}{\linewidth}{
@{}>{\scriptsize\raggedright\arraybackslash}l
*{3}{>{\scriptsize\centering\arraybackslash}X}
@{}}
\toprule
\textbf{Method}
& \textbf{GPQA-D}
& \textbf{MMLU-Pro}
& \textbf{Avg.} \\
\midrule

\multicolumn{4}{@{}l}{\emph{\textbf{Qwen3-8B-Base}}} \\
DAPO & 43.43 & 63.97 & 58.87 \\
\rowcolor{cyan!10}
\textbf{DelTA} & \textbf{50.00} & \textbf{66.47} & \textbf{62.38} \\
\midrule

\multicolumn{4}{@{}l}{\emph{\textbf{Qwen3-14B-Base}}} \\
DAPO & 54.55 & 70.80 & 66.77 \\
\rowcolor{cyan!10}
\textbf{DelTA} & \textbf{56.67} & \textbf{72.27} & \textbf{68.40} \\
\bottomrule
\end{tabularx}

\vspace{-10pt}
\end{wraptable}
To examine whether DelTA generalizes beyond the in-domain mathematical reasoning benchmarks, we further evaluate DAPO and DelTA on two out-of-domain benchmarks: GPQA-Diamond and MMLU-Pro. Since MMLU-Pro contains a large number of questions, we randomly sample 600 questions for evaluation. For each question, we sample 5 responses and report Avg@5. We also report a question-count weighted average across the two benchmarks.

As shown in Table~\ref{tab:results_on_ood}, DelTA consistently improves over DAPO on both OOD benchmarks and both backbone sizes. On Qwen3-8B-Base, DelTA improves the weighted average from \(58.87\) to \(62.38\). On Qwen3-14B-Base, DelTA further improves the weighted average from \(66.77\) to \(68.40\). These results suggest that DelTA does not merely overfit to the main mathematical evaluation suite, but also transfers to broader scientific and general knowledge reasoning tasks.

\subsection{Token weight analysis}
\label{app:token-weight-analysis}

\begin{wrapfigure}[17]{r}{0.3\linewidth}
    \vspace{-3mm}
    \centering

    \includegraphics[width=\linewidth]{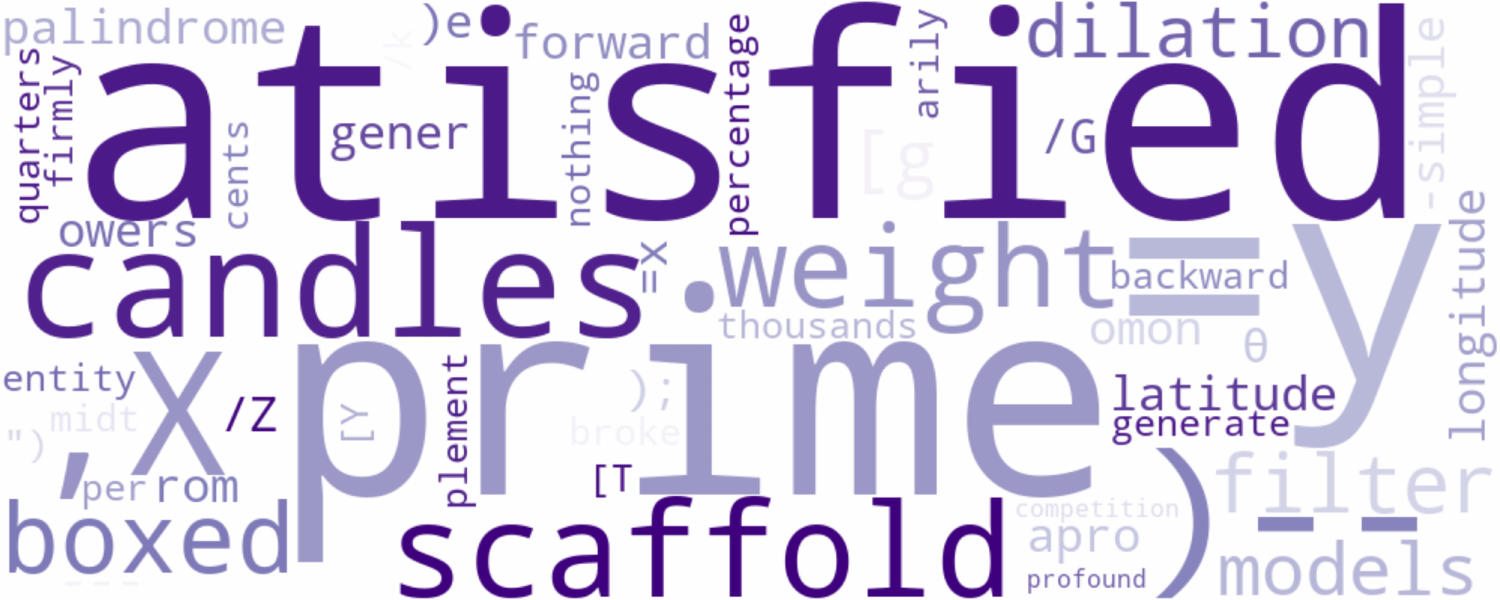}
    \vspace{-1mm}
    
    {\footnotesize \textbf{(a)} High-weight tokens\par}

    \vspace{1.5mm}

    \includegraphics[width=\linewidth]{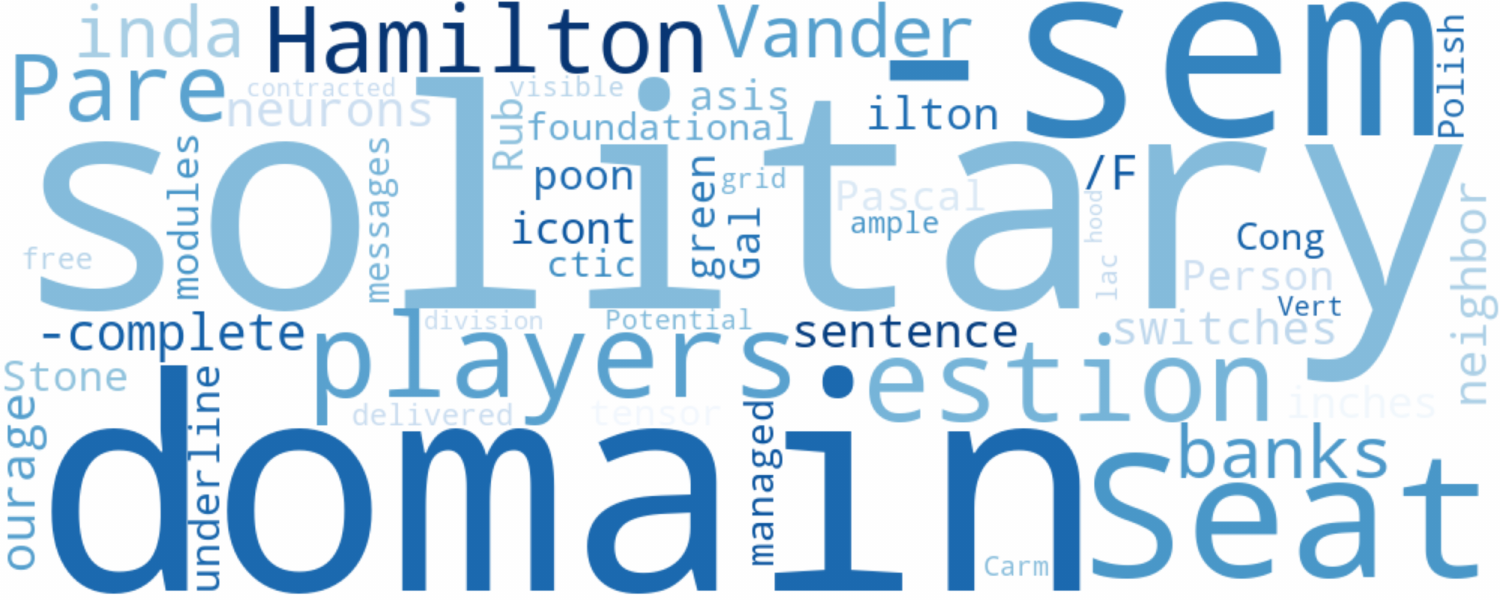}
    \vspace{-1mm}
    
    {\footnotesize \textbf{(b)} Low-weight tokens\par}

    \vspace{-1mm}
    \caption{Token clouds of high-weight and low-weight tokens.}
    \label{fig:token_cloud}
    \vspace{-4mm}
\end{wrapfigure}
To understand what DelTA emphasizes, we visualize high- and low-weight token clouds in Figure~\ref{fig:token_cloud}. The visualization is based on selected generated tokens collected during training at a generation scale of roughly \(10^8\) tokens. For each token type, we compute its average DelTA coefficient over its occurrences; larger tokens indicate higher average coefficients in the high-weight cloud and lower average coefficients in the low-weight cloud. This context-agnostic visualization is qualitative, since DelTA assigns coefficients to tokens based on their gradient vectors in specific rollout contexts rather than to token strings in isolation. The high-weight cloud contains several tokens related to transformations and verification, such as \texttt{scaffold}, \texttt{prime}, \texttt{=y}, \texttt{forward}, and \texttt{backward}. In contrast, the low-weight cloud is dominated by more background-like or entity-specific tokens, such as \texttt{Seat}, \texttt{domain}, \texttt{players}, \texttt{Vander}, and \texttt{Hamilton}.

This pattern is consistent with DelTA's design. Rather than reflecting semantic importance alone, \(\lambda_{i,t}\) reflects whether the gradient vector of a token is more representative of its own advantage side than of the opposite side. DelTA therefore tends to assign larger coefficients to more discriminative reasoning-related tokens and smaller coefficients to shared or less informative background tokens.

\end{document}